\pdfoutput=1

\documentclass[11pt]{article}

\usepackage[preprint]{acl}

\usepackage{times}
\usepackage{latexsym}

\usepackage[T1]{fontenc}

\usepackage[utf8]{inputenc}

\usepackage{microtype}

\usepackage{inconsolata}

\usepackage{xcolor}
\usepackage{graphicx}
\usepackage{booktabs}
\usepackage{amsmath}
\usepackage{makecell}
\usepackage{colortbl}
\usepackage{array}
\usepackage{longtable}
\title{Natural Language Processing in the Patent Domain: A Survey}

\author{Lekang Jiang \and Stephan M. Goetz \\
        University of Cambridge \\
        \texttt{\{lj408, smg84\}@cam.ac.uk}}

\begin{document}
\maketitle
\begin{abstract}

Patents, which encapsulate crucial technical and legal information in text form and referenced drawings, present a rich domain for natural language processing (NLP) applications. As NLP technologies evolve, large language models (LLMs) have demonstrated outstanding capabilities in general text processing and generation tasks. However, the application of LLMs in the patent domain remains under-explored and under-developed due to the complexity of patents, particularly their language and legal framework. Understanding the unique characteristics of patent documents and related research in the patent domain becomes essential for researchers to apply these tools effectively. 
Therefore, this paper aims to equip NLP researchers with the essential knowledge to navigate this complex domain efficiently.
We introduce the relevant fundamental aspects of patents to provide solid background information. 
In addition, we systematically break down the structural and linguistic characteristics unique to patents and map out how NLP can be leveraged for patent analysis and generation. Moreover, we demonstrate the spectrum of text-based and multimodal patent-related tasks, including nine patent analysis and four patent generation tasks.\footnote{This is an updated version of the preprint. The final version is published in \textit{Artificial Intelligence Review}. Please refer to and cite the final version: \url{https://doi.org/10.1007/s10462-025-11168-z}.}

\end{abstract}

\section{Introduction}

Patents, a form of intellectual property (IP), grant the holder temporary rights to suppress competing use of an invention in exchange for a complete disclosure of the invention. It was once established to promote and/or control technical innovation and progress \citep{Frumkin1947}. The surge in global patent applications and the rapid technological progress pose formidable challenges to patent offices and related practitioners \citep{krestel2021survey}. 
These challenges overwhelm traditional manual methods of patent drafting and analysis. Consequently, there is a significant need for advanced computational techniques to automate patent-related tasks. Such automation not only enhances the efficiency of patent and IP management but also facilitates the extraction of valuable information from this extensive knowledge base \citep{abbas2014literature}.

Researchers have investigated machine learning (ML) and natural language processing (NLP) methods for the patent field with highly technical and legal texts \citep{krestel2021survey}. In addition, the recent large language models (LLMs) have demonstrated outstanding capabilities across a wide range of general domain tasks \citep{zhao2023survey,min2023recent}. Moreover, the expansion of the latest general LLMs with a graphical component to form multimodal models \citep{huang2024large} may further enhance the capabilities in processing patents, which include text and drawings. These models are promising to become valuable tools in managing and drafting patent literature, the crucial resource that documents technological advances. 

However, compared to the significant success of LLMs in the general domain, the application of LLMs in patent-related tasks remains under-explored due to the texts' and the field's complexity. NLP and multimodal model researchers need to deeply understand the unique characteristics of patent documents to develop useful models for the patent field. Therefore, we aim to equip researchers with the essential knowledge by presenting this highly auspicious but widely neglected field to the NLP community.

\begin{figure*}[!htbp]
    \centering
    \includegraphics[width=\textwidth]{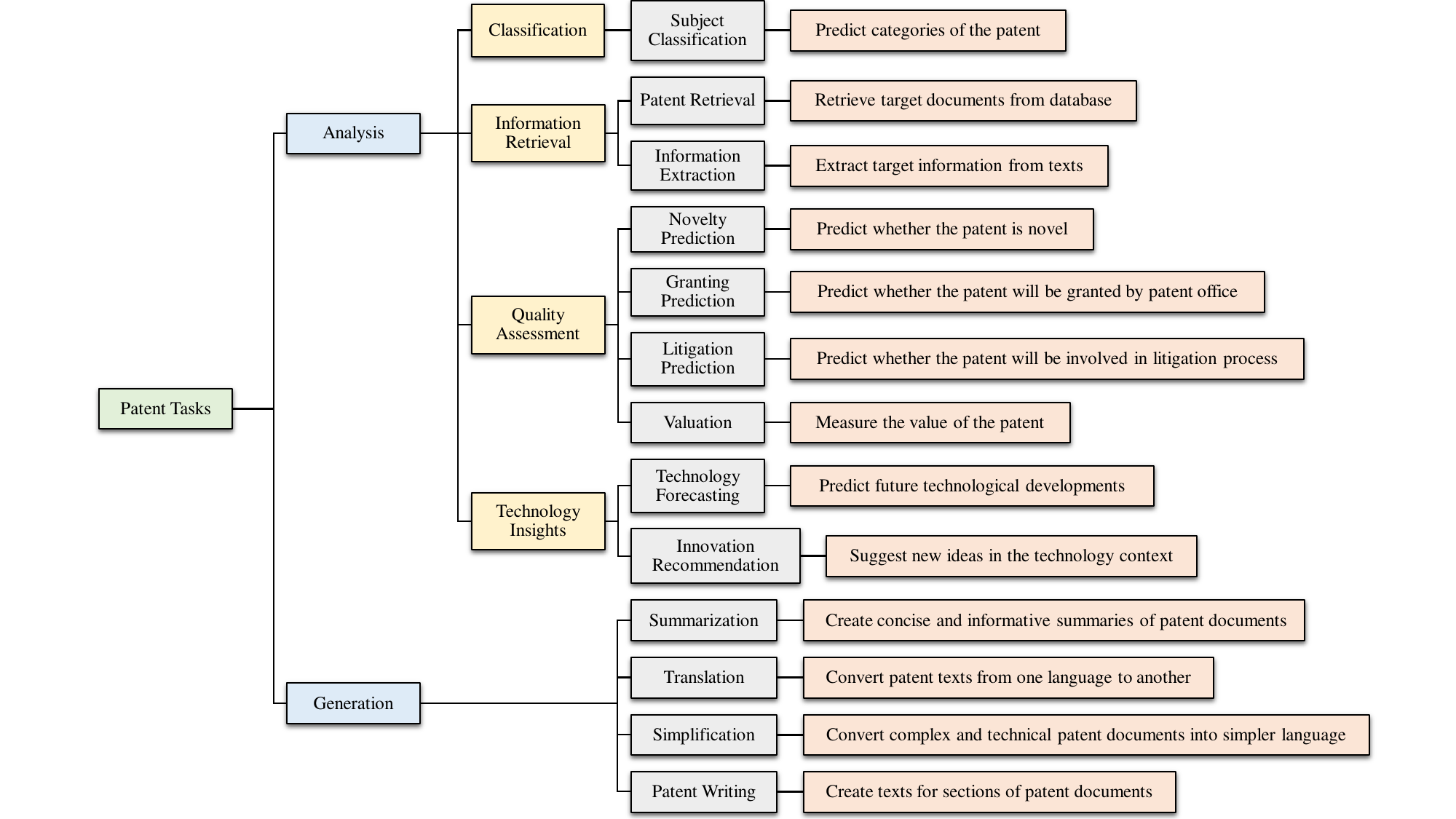}
    \caption[]{Patent-related tasks }
    \label{fig:task}
\end{figure*}

Previous surveys reported the early stages of smart and automated methods for patent analysis \citep{abbas2014literature}, first deep learning methods, which opened a wider range of still simpler patent tasks \citep{krestel2021survey}, or specific individual patent tasks, such as patent retrieval \citep{shalaby2019patent}. The recent advancements in language and multimodal models were unforeseen, particularly the performance boost when models are massively scaled up \citep{kaplan2020scaling}. Accordingly, we specifically delineate a survey of popular methodologies for patent documents with a special focus on the most recent and evolving techniques. We have included the two applications of patent analysis and generative patent tasks (Fig.~\ref{fig:task}). Whereas analysis focuses on understanding and usage of individual patent documents or a group of patents, generation tasks aim at automatically generating patent texts.

We provide a systematic survey of NLP applications in the patent domain, including fundamental concepts, insights on patent texts, development trends, datasets, tasks, and future research directions to serve as a reference for both novices and experts. Specifically, we cover the following topics:

\begin{figure*}[!t]
    \centering
    \includegraphics[width=\textwidth]{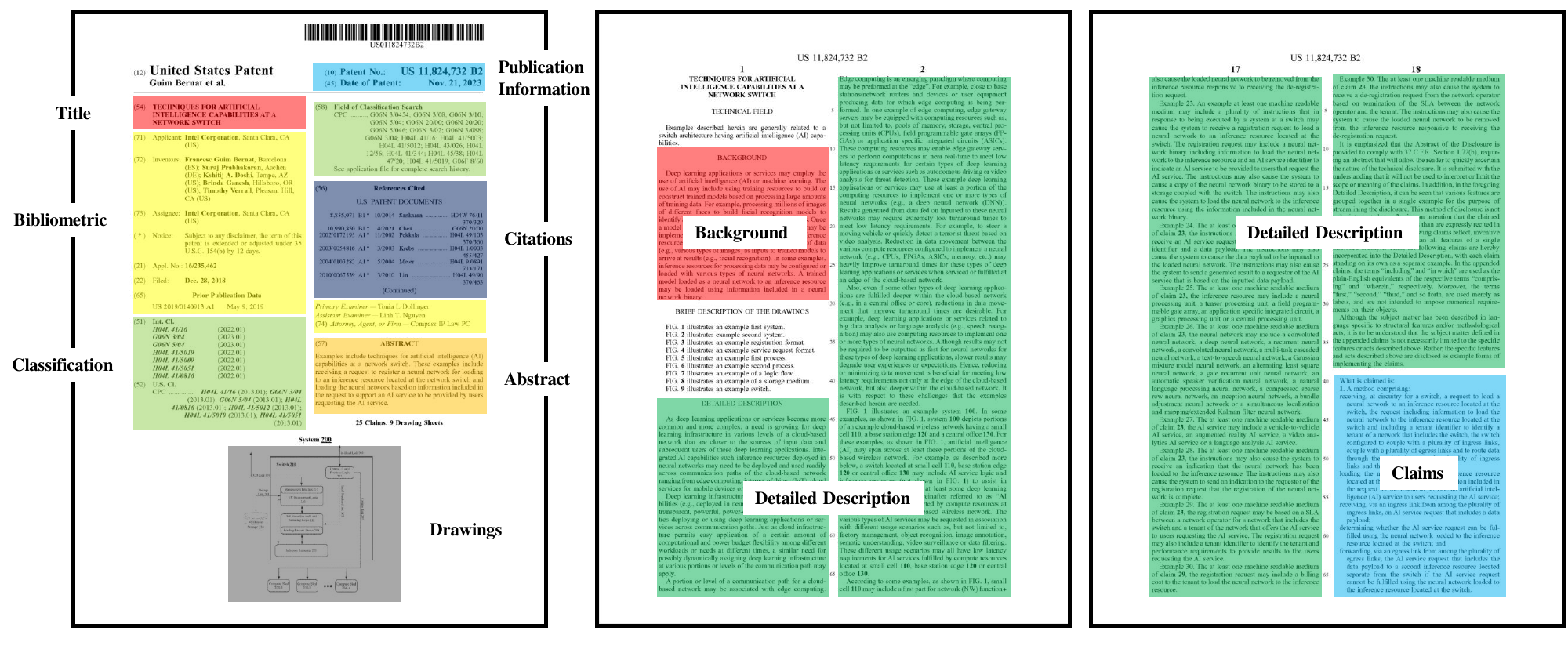}
    \caption[]{An example of granted patent document (US Patent 11,824,732 B2). This is a US patent, and other countries have their own patent systems with largely similar requirements. }
    \label{fig:document}
\end{figure*}

\begin{enumerate}

\item We provide an introduction to the fundamental aspects of patents in Section~\ref{background}, including the composition of patents and the patent life-cycle. This section is particularly for those readers still unfamiliar with it and a refresher for others. 

\item We analyze the unique structural and linguistic characteristics of patent texts with language processing and multimodal techniques to process these texts in Section~\ref{s:insight}. Readers will readily understand the challenges of automated patent processing and the development trends of NLP in the patent field.

\item We present a detailed synthesis of patent data sources, alongside tailored datasets specifically designed for different patent tasks, in Section~\ref{data}. With the collection of these ready-to-use datasets, we aim to eliminate the extensive time and effort for data preparation in the patent domain.

\item We evaluate nine patent analysis tasks (subject classification, patent retrieval, information extraction, novelty prediction, granting prediction, litigation prediction, valuation, technology forecasting, and innovation recommendation) in Section~\ref{analysistask} and four generation tasks (summarization, translation, simplification, and patent writing) in Section~\ref{generationtask}. Specifically, we systematically demonstrate task definitions and relevant methodologies in detail. We show a comprehensive yet accessible insight into this area, which should enable readers to grasp the nuances of the field more effectively.

\item We identify current challenges and point out potential future research directions in Section~\ref{futurework}. By highlighting these areas, we hope to encourage further research and development in automated patent tasks to stimulate more efficient and effective methods in the future.

\end{enumerate}

\section{Brief Background}
\label{background} 

\subsection{Patent Document}
\label{document}
Patent documents are central elements for the protection of intellectual property and also document inventions. Patents require applicants and/or inventors to publicly disclose their inventions in detail to secure exclusive rights and obtain benefits in return. Patent documents describe new inventions and delineate the scope of patent rights granted to patent holders. These documents are key parts of the patenting process and are publicly accessible typically 18 months after the application or the first filing date. The format and content can vary by jurisdiction but normally include the following elements. Fig.~\ref{fig:document} displays an example patent document.

\noindent \textbf{Publication information} includes the file number and date of patent (application) publication.

\noindent\textbf{Title} is the concise description of the invention.

\noindent\textbf{Bibliometric information} includes details about applicants, inventors, assignees, examiners, attorneys, etc. 

\noindent\textbf{Patent classification} code defines the category of the patent. We introduce detailed information on patent classification schemes in Section~\ref{class_scheme}.

\noindent\textbf{Citations} are lists of prior arts and other patents referenced in the document or by examiners.

\noindent\textbf{Abstract} is a brief summary of the invention and its purpose.

\noindent\textbf{Background} contains basic information on the field of the invention and is supposed to list and appreciate the prior art, particularly in the patent literature.

\noindent\textbf{Detailed description} provides comprehensive details about the invention and specific embodiments, typically discussing the drawings.

\noindent\textbf{Claims} define the legal scope of the patent. Each claim is a single sentence, which describes the invention in specific features that make the invention novel and not easily derivable (obvious) from the state of the art (described by any source, not only patent documents).

\begin{figure}[!t]
    \centering
    \includegraphics[width=.4\textwidth]{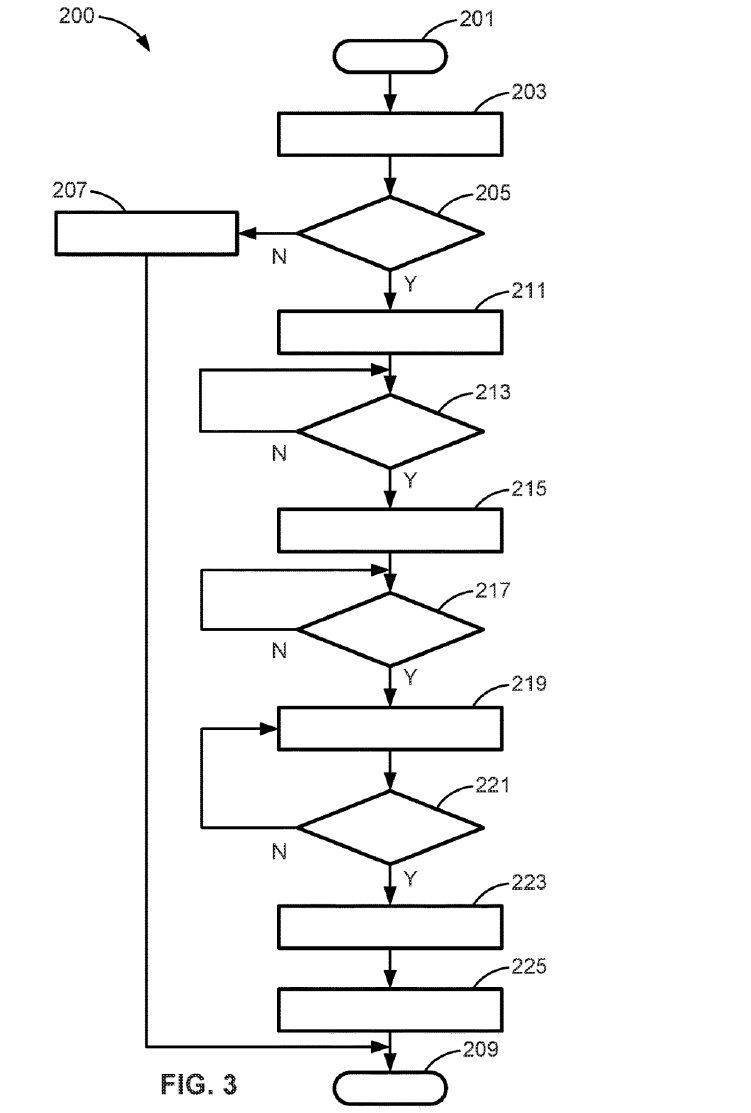}
    \caption[]{An example drawing from patent US 10,854,933. Many figures in patents are generic without the corresponding description. Drawings tend to be less generic in patents on pharmaceuticals or mechanics, often consisting of graphs, images, or models. Some drawings are also of poor resolution, pixelation, or low quality. The reference numbers indicate specific elements introduced in the patent description. The reference numbers have to be named by the same term consistently throughout the patent application, which can also substantially deviate from conventions in the field. The reference numbers are typically used for invention features and listed in the claims.}
    \label{fig:drawing}
\end{figure}

\noindent\textbf{Drawings} are visual representations of the invention, disclosing important aspects of the invention as well as embodiments to support the textual description. Fig.~\ref{fig:drawing} shows an example patent drawing. Apart from patent texts, researchers also use patent drawings for patent analysis, which is introduced in Appendix~\ref{drawings}. 

\subsection{Patent Life Cycle}
\label{lifecycle}

\begin{figure}[!t]
    \centering
    \includegraphics[width=.49\textwidth]{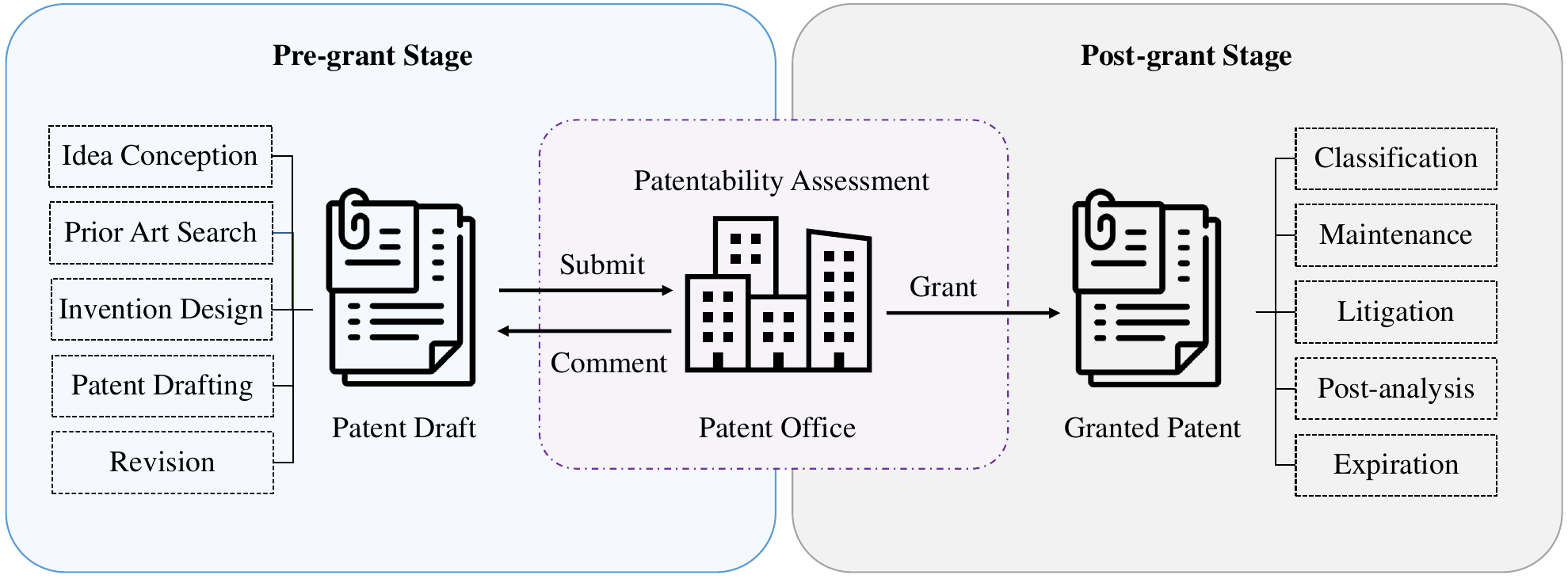}
    \caption[]{Patent life-cycle from pre-grant to post-grant stage }
    \label{fig:application}
\end{figure}

The patent life cycle encompasses several stages, from the initial conception of an invention to its eventual expiration. It can be broadly divided into pre-grant and post-grant phases (Fig.~\ref{fig:application}). 

\noindent\textbf{Pre-Grant Stage. } In the beginning, inventors conceptualize, design, and develop their invention. If inventors hope to obtain patent protection for their invention, they need to apply to the patent office. To ensure novelty and inventiveness (non-obviousness), inventors may preventively search for existing patents and public disclosures to avoid unnecessary cost and effort in case of prior disclosures. Additionally, patent documents need drafting to describe the invention in detail, including its specifications, claims, abstract, and accompanying drawings. The drafting process typically requires the expertise of a patent professional, such as a patent engineer, attorney, or agent. After the patent application is submitted to the patent office, examiners at the office will screen and evaluate the application for compliance with formal, legal, and technical requirements. This examination typically involves correspondence of the examiner with the inventor or their representation, where clarifications, amendments, or arguments are submitted.\footnote{Amendments must not introduce novel features, i.e., inventive content, to avoid losing the original date stamp or the entire file.} Notably,  an assessment will determine if the invention meets the criteria for patentability, including novelty, inventiveness (non-obviousness), and typically less strictly commercial utility. The requirement of utility may be less strict as offices may see that as the applicant's problem, except for certain cases. The substantive examination compares the invention with similar documents from the patent literature or any other public source dated earlier, which were found in an initial search by the office. Upon examination and potential resolution of any objections, the patent is either granted---rarely in its original, more frequently in a restricted form based on the identified prior art---or rejected. If the original priority date should be maintained, the invention must not be extended during the examination.\footnote{Such additional material should not be added casually but would require proper announcement with a new filing that typically claims the priority of the base filing for all aspects already described in the latter. Otherwise, the inadmissibly extended file may be rejected or later challenged in the granted stage.}

\noindent\textbf{Post-Grant Stage.} The granted patent is published, disclosing the details of the invention to the public potentially accompanying a previous application publication. As a remnant of previous pre-computer and pre-AI-search times, the offices classify patents by field of technology for easier search and management. Maintenance fees need to be paid regularly to keep the patent in force. The patent owner can enforce patent rights through legal action when infringement occurs. In addition, third parties can question the validity of patents, and patent owners would need to defend the challenges. Furthermore, companies can analyze patents for technology insights and derive strategies. Finally, the invention enters the public domain after the expiration of patents, allowing anyone to use it without infringement if no other rights cover those aspects.

\subsection{Related Surveys}
\label{related}
\noindent\textbf{Patent Analysis. } Various surveys in the past summarized certain aspects of patent analysis from a knowledge or procedural perspective. The work of \citet{abbas2014literature} represents early research on patent analysis. These early methods included text mining and visualization approaches, which paved the way for future research. Deep learning for knowledge and patent management started more recently. \citet{krestel2021survey} identified eight types of sub-tasks attractive for deep learning methods, specifically supporting tasks, patent classification, patent retrieval, patent valuation, technology forecasting, patent text generation, litigation analysis, and computer vision tasks. Some surveys reviewed specific topics of patent analysis. For example, \citet{shalaby2019patent} surveyed patent retrieval, i.e., the search for relevant patent documents, which may appear similar to a web search but has substantially different objectives as well as constraints and has a legally-defined formal character. A concurrent survey introduced the latest development of patent retrieval and posed some challenges for future work \citep{ali2024innovating}. In addition, \citet{balsmeier2018machine} highlighted how machine learning and NLP tools could be applied to patent data for tasks such as innovation assessment. It focused more on methodological advancement in patent data analysis that complements the broader survey of tools and models used in NLP for patent processing.

\noindent\textbf{Patent Usage. } The patent literature records major technological progress and constitutes a knowledge database. It is well known that the translation of methods from one domain to another or the combination of technology from various fields can lead to major innovation. Thus, the contents of the patent literature appear highly attractive for systematic analysis and innovation design. However, the language of modern patents has substantially evolved and diverged from normal technical writing. Recent patent documents are typically hardly digestible for the normal reader and also contain deceptive elements added for increasing the scope or to camouflage important aspects. Appropriate data science techniques need to consider such aspects to mine patent databases for engineering design \citep{jiang2022patent}. Data science applied to this body of codified technical knowledge cannot only inform design theory and methodology but also form design tools and strategies. Particularly in early stages of innovation, language processing can help with ideation, trend forecasting, or problem--solution matching \citep{just2024natural}.

\noindent\textbf{Patent-Related Language Generation. } Whereas the above-listed techniques harvest the existing patent literature for external needs, the strong text-based patent field suggests the use of language models in a generative way. Language processing techniques can, for instance, translate the peculiar patent language to more understandable technical texts, summarize texts, or generate patent texts based on prompts \citep{casola2022summarization}. The recent rapid development of generative language models, especially large language models (LLMs) \citep{zhao2023survey} may stimulate more patent-related generation tasks.

\noindent\textbf{Related Topics. } A variety of other tasks and applications also have close ties to the patent domain. For example, intellectual property (IP) includes copyrights, trademarks, designs, and a number of more special rights beyond patents.  The combined intellectual property literature allows concordant knowledge management, technology management, economic value estimation, and information extraction \citep{aristodemou2018state}. The patent field is closely related to the general legal domain, because they share procedural aspects and the precision of language. Likewise, most patent professionals have substantial legal training or a law degree, and their work often involves legal aspects. Accordingly, NLP techniques in the more general field of law may influence patent-related techniques in the future \citep{katz2023natural}.

\section{Insights into Patent Texts}
\label{s:insight}

Patent text can differ from normal text in multiple aspects, which stimulates a variety of research and entails challenges for the field. 

\subsection{Long Context }
Most research so far has focused on short texts, such as patent abstracts and claims. However, titles and abstracts of patents can be surprisingly generic. Therefore, using patent descriptions that provide comprehensive details and specific embodiment of the invention for patent analysis is highly important, which is neglected by most of the current research. A possible reason is that previous language models cannot handle such long inputs. According to a recently proposed patent dataset \citep{suzgun2022harvard}, the average number of tokens of a patent description exceeds 11,000, which is longer than the context limit of many previous language models. For example, Llama-2 supports the context length of 4,000 tokens \citep{touvron2023llama}. The reason for the limited context length of many language models is the rapid growth of computational complexity associated with self-attention. It grows because the number of attention relationships increases with the context length. As self-attention often considers every pair of tokens, the computational complexity can grow with the context length squared ($\mathcal{O}(n^2)$). The long context of patent descriptions causes critical challenges for patent analysis. 

Notably, researchers have investigated increasing or otherwise handling the context length of LLMs \citep{chen2023clex, jiang2023mistral, xiong2023effective}. \citet{xiong2023effective}, for example, introduced a series of long-context LLMs, which can support the context windows up to 32,768 tokens. A mix of reduced short-range and long-range attention instead of exhaustive attention can reduce the computational and memory burden \citep{kovaleva2019revealingdarksecretsbert,beltagy2020longformer}. Moreover, the latest very large models, such as GPT-4\footnote{\url{https://platform.openai.com/docs/models/gpt-4-and-gpt-4-turbo}} \citep{2023arXiv230308774O} and Llama-3.1\footnote{\url{https://ai.meta.com/blog/meta-llama-3-1/?_fb_noscript=1}} \citep{dubey2024llama} even naturally support a context length of 128,000 tokens, which appears promising to process the long patent descriptions. Other LLMs also showed a trend towards long context capabilities, such as Falcon-180B \citep{falcon}, Gemini 1.5\footnote{\url{https://blog.google/technology/ai/google-gemini-next-generation-model-february-2024/}} \citep{reid2024gemini} and Claude 3.5\footnote{\url{https://www.anthropic.com/news/claude-3-5-sonnet}} \citep{claude3_2024}.

\subsection{Technical Language} 
The patent language is highly technical and artificial, including specialized terminology, legal phrases, and sometimes newly coined terms to describe new concepts that may not yet have been widely recognized. Patents regularly define their own terms, which may substantially deviate from everyday language usage and the technical field. Such self-defined terms are often highly artificial and not likely to occur in other relevant documents or even in any dictionary. 

Hence, the technical language causes significant challenges for general LLMs for patent analysis, which are trained on normal texts and a large share of more colloquial language. Thus, LLMs may not capture the patent context information effectively, because the important technical terms can be completely new to the LLMs or have different meanings from its pre-training corpora. Embeddings based on distance metrics for synonymity or semantic relationships of terms may not work if terms are defined contrary to their normal use or are entirely new.

\subsection{Precision Requirement }
The precision requirement and information density of patent texts are higher than in everyday language.\footnote{It is worth mentioning that the high importance of legal and formal aspects can tip the balance away from the sciences. Prophetic examples or experiments in patent documents describe anticipated results or theoretical experiments that have not been physically conducted. These are hypothetical scenarios intended to demonstrate the scope of an invention or to anticipate and complicate any competing patent filing for competitors with wild speculations. While many offices only require a basic feasibility of the invention per the claims, such documents can receive incorrect citations and let these prophecies or speculations appear as factual, which can lead to misinformation \citep{freilich2019prophetic}.} The patent language focuses more on precision and accuracy than on readability. Patent texts must be precise and meticulously described to ensure the patent is both defensible and enforceable. Such precision requirement typically leads to high repetitiveness in both terminology and structure of sentences, paragraphs, and sections. Furthermore, sentences are often overburdened because they use relative or adverbial clauses to include specifications for precision or add examples for a wider scope. Additionally, each terminology must be used consistently throughout the document. That means a technical term must not be replaced by other words unless the patent explicitly states that both are identical. In contrast, everyday texts and academic literature tend to vary and paraphrase the wording for better readability. In addition, the patent claim is carefully crafted to define the precise scope and boundaries of the invention's protection, ensuring that the patent can withstand legal scrutiny. 

The precision requirement of patent texts complicates the patent generation tasks because LLMs are likely to generate slightly different words or phrases. Due to the requirement for large quantities of data, most pre-training corpora for LLMs tend to be colloquial and relatively informal. Another smaller portion comes from literary texts, which may be of higher quality but typically prioritize style and linguistic originality over precision and accuracy. 

\subsection{LLMs for Patent Processing}

Patent texts are distinct from everyday texts with respect to long context length, in-depth technical complexity, and high precision to ensure the patent can be granted, defended, and enforced. 
This type of language often necessitates specific training and experience in patents for accurate reading and interpretation. 
However, human readers usually struggle with certain aspects of patent texts, such as ambiguous names, phrases that conflict with their prior understanding, or numerous terms that the description redefines within the context of the specific patent itself. The dense, specialized terminology and unconventional syntactic structures common in patents often pose significant barriers to comprehension, even for those with experience in technical fields.
In contrast, LLMs trained on datasets specifically tailored to patent language are theoretically well-equipped to handle these challenges. They are designed to process complex syntactic structures, manage long-range dependencies, and incorporate newly coined terms or domain-specific jargon that diverges from everyday language. By leveraging the specialized vocabulary of experts in the field, these models can navigate the nuanced requirements of patent texts with a level of consistency and scope that may exceed human capabilities. Therefore, the recent advancements in LLMs \citep{zhao2023survey} for generative tasks and language processing appear ideally suited for the unique characteristics of patent literature. They promise large benefits in areas, such as patent drafting, prior art search, and examination.

Despite the apparent fundamental compatibility of LLMs for knowledge extraction and language processing, the application of LLMs in the patent domain remains underdeveloped and not yet highly prominent. Previous studies used word embeddings (e.g., Word2Vec \citep{mikolov2013efficient}) and deep learning models (e.g., LSTM \citep{hochreiter1997long}) for patent analysis tasks. As transformers \citep{vaswani2017attention} showed significant potential in text processing, researchers started to develop transformer-based language models, such as BERT \citep{devlin2018bert} and GPT \citep{radford2018improving}.  
The recent large-sized models with outstanding capabilities have not been extensively investigated in the patent field. 
Some representative general LLMs that are worth exploring include the Llama-3 family \citep{dubey2024llama}, Mistral \citep{jiang2023mistral}, Mixtral \citep{jiang2024mixtral}, GPT-4 \citep{2023arXiv230308774O}, Claude 3 \citep{claude3_2024}, DeepSeek-V3 \citep{liu2024deepseek}, and Gemini 1.5 \citep{reid2024gemini}. Researchers have also developed patent-specific LLMs, such as PatentGPT-J \citep{lee2023evaluating} and PatentGPT \citep{bai2024patentgpt}. However, PatentGPT-J has shown limited performance in patent text generation tasks \citep{jiang2024can}, and PatentGPT, though promising, is a recent development that is not yet publicly available. Significant work related to patent-specific LLMs remains to be done. Moreover, since patents represent a type of legal document, law-specific LLMs are also worth investigating, such as SaulLM \citep{colombo2024saullm}. 

Previous research efforts have included patent analysis, data extraction, and automation of procedures. However, the lack of benchmark tests, such as reference datasets and established metrics, hinders performance evaluation and comparison across different methods. The effectiveness of LLMs depends on the quality of training data. To this end, we have compiled sources and databases of patents with curated datasets tailored for various patent-related tasks in Section~\ref{data}. Although patent offices have released raw documents for years, publicly available datasets for specific tasks remain scarce. Numerous studies continue to use closed-source data for training and evaluation. Furthermore, patent offices do not provide pre-processed data or broad access to well-structured documents from the process around patents. Although in many countries, patent documents as the disclosure of inventions are considered public domain, they offer only the manual review of individual documents.\footnote{This aspect may reflect an important discussion to be led by the field and society. As in other domains, there may be a missing or unclear consensus on which data can be used for machine learning. In stark contrast to other domains, where even books and copyrighted materials may be processed in large numbers, sometimes to create new works that compete with the originals, the patent system is a bit different. The entire modern patent system was developed and is justified as a social contract, trading invention disclosure for temporary exclusion from competition. The descriptions are often public domain and related correspondence with patent offices is made public to ensure transparency.  Thus, imposing paywalls would typically violate the mission of patent offices. Admittedly, big-data methods and artificial intelligence were not considerations during the establishment of these legal frameworks.  Patent offices may still restrict or complicate larger-scale data access to parts or all contents to avoid commercial exploitation. Some patent offices offer costly paid database access as a stream of revenue, which can be prohibitively expensive for academic research. Important context to know is that in many places patent offices need to fund their own operating expenses \citep{USPTOBudget2024, EPOFinancialStatement2022, Pottelsberghe2009PatentCost, frakes2014failed}.  Therefore, society—which the patent offices are ultimately meant to serve—must reach a consensus on how to manage data access for commercial versus academic purposes, potentially leading to both proprietary and open-source models.} The pre-processing steps to formulate structured patent datasets generally involve: segmenting patent documents into clearly defined sections to enhance downstream tasks, such as abstract, claims, description, etc; and rectifying irregularity issues, such as missing fields, erroneous characters, or formatting issues. 

Previous research prominently focuses on short text parts of patents, such as titles and abstracts. However, these texts are typically highly generic and include the least specific texts with little information about the actual invention. These texts also take little time during drafting, making the automated creation of these parts less helpful. By contrast, patent descriptions need to include all details of an invention, and patent claims clearly define the legal boundaries of the invention's protection, the so-called scope. Since these texts are longer and contain much more useful information, they are worth more attention for both patent analysis and generation.

\subsection{Multimodal Techniques}

A patent is not merely a text document but can include drawings, i.e., visual components. Thus, multimodal methods such as CLIP \citep{radford2021learning} and Vision Transformers \citep{dosovitskiy2020image} may unlock this potential. Multimodal methods in patent processing integrate diverse data types, such as text, images, and quantitative information, to enhance tasks such as classification and retrieval. The combination of their complementary strengths of different modalities may lead to more comprehensive and accurate results. 
\citet{lee2022multimodal} for instance introduced a multimodal deep-learning model that combined textual content with quantitative patent information, which resulted in improved performance for patent classification tasks. Additionally, multimodal approaches that integrate textual descriptions with visual content have shown promise in enhancing patent retrieval \citep{lo2024large}. Furthermore, \citet{lin2023measuring} proposed multimodal methods to extract structural and visual features to effectively measure patent similarity. Such multimodal similarity detection may improve the efficiency of patent examination.
However, as illustrated in Fig.~\ref{fig:drawing}, many drawings in patents are generic without the corresponding description, and some drawings suffer from poor resolution, pixelation, or low quality. This may be one of the reasons why current multimodal methods in patent processing are scarce. Moreover, textual elements, especially the claims and descriptions, are the primary carriers of legally binding information in patents, particularly features. While drawings provide valuable support and document embodiments, the text is essential for defining the scope, novelty, and context of each invention. Consequently, the survey particularly examines NLP approaches and includes representative multimodal methods where available.

\section{Data}
\label{data}
 
\subsection{Data Sources}
\label{datasource}
Patent applications are submitted to and granted by patent offices. To stimulate innovation and serve the society, patent offices provide detailed information about existing patents, patent applications, and the legal status of patents---previously on paper, nowadays online. Large patent offices include the United States Patent and Trademark Office (USPTO) and the European Patent Office (EPO). These offices also provide access to download bulk datasets or tools to explore and analyze patent data. For example, PatentsView is a platform developed by the USPTO, providing accessible and user-friendly interfaces to explore US patent data, with various tools for visualization and analysis. Apart from patent offices, there are searchable databases that contain patents from multiple countries and offices, such as Google Patents. We list a broad range of patent offices and databases in Table~\ref{patentsource}.

\begin{table*}[!htbp]
\centering
\begin{tabular}{p{4cm}p{11cm}}
\toprule
Patent Office \& Database       & \multicolumn{1}{c}{Brief Description}       \\ 
\midrule
USPTO               & 
United States Patent and Trademark Office is a comprehensive source for US patents and applications. \url{https://www.uspto.gov/}     \\
EPO  & 
European Patent Office offers a variety of tools, such as Espacenet, for searching European and also foreign patents, which the tools can link to patent families. \url{https://www.epo.org/en}                                   \\
WIPO  & 
World Intellectual Property Organization includes global patent documents and applications from participating national and regional patent offices. \url{https://www.wipo.int/portal/en/index.html}    \\
CNIPA / SIPO   & 
State Intellectual Property Office, formerly China National Intellectual Property Administration, provides a database for searching Chinese patents. \url{http://english.cnipa.gov.cn/} \\
JPO      & 
Japan Patent Office provides information on Japanese patents and utility models. \url{https://www.jpo.go.jp/e/}                                 \\
UK IPO  &
United Kingdom Intellectual Property Office gives access to UK, Jersey, and Guernsey patents, . \url{https://www.ipo.gov.uk/}     \\
KIPO  &
Korea Intellectual Property Office includes the Korean Patent System and Utility Model System. \url{https://www.kipo.go.kr/}     \\
Auspat  & 
The Australian patent database. \url{https://www.ipaustralia.gov.au/}    \\
Google Patents  & 
This is a free and searchable database of patents and patent applications from multiple countries. \url{https://patents.google.com/}\\
PatentsView  & 
A platform developed by the USPTO, providing interfaces for exploring U.S. patent data. \url{https://patentsview.org/}   \\
Espacenet  &
Espacenet allows free access to over 140 million patent documents provided by EPO. \url{https://worldwide.espacenet.com/} \\
Patentscope   & 
Patentscope belongs to WIPO, allowing searches in international patent collections. \url{https://www.wipo.int/patentscope/en/}  \\
Global Dossier   & 
Global Dossier is developed and maintained by a collaboration of USPTO, EPO, JPO, KIPO, and SIPO to provide a shared database and access point to IP registers and the corresponding procedural files. \url{https://globaldossier.uspto.gov}  \\
Lens  & 
Lens integrates scholarly and patent knowledge,  providing patent search and analysis tools. \url{https://www.lens.org/}       \\ \bottomrule
\end{tabular}
\caption{Patent offices and databases with brief descriptions. }
\label{patentsource}
\end{table*}

\subsection{Curated Data Collections}
\label{curateddata}

\begin{table*}[!htbp]
\centering
\resizebox{\textwidth}{!}{
\begin{tabular}{lllcp{8cm}}
\toprule
Dataset                                                                                                        & Source & \multicolumn{1}{c}{Task}                                                             & Size                                                              & \multicolumn{1}{c}{Brief description}                                                                                                                                                                                                \\ \midrule
USPTO-2M \citep{li2018deeppatent}                                                                               & USPTO  & Classification                                                   & 2M                                                                & Utility patent documents classified into 637 classes. Original link: \url{http://mleg.cse.sc.edu/DeepPatent/}                                                                                                                        \\
USPTO-3M \citep{lee2020patent}                                                                                  & USPTO  & Classification                                                   & 3M                                                                & Collected from Google Patents Public Datasets using SQL. Data information at: \url{https://doi.org/10.1016/j.wpi.2020.101965}                                                                                                        \\
WIPO-Alpha \citep{fall2003automated}                                                                            & WIPO   & Classification                                                   & 75K                                                               & Documents classified into 114 classes and 451 subclasses. Original link: \url{https://www.wipo.int/ibis/datasets}                                                                                                                    \\
\begin{tabular}[c]{@{}l@{}}Crop Protection\\ Industry dataset \citep{christofidellis2023automated}\end{tabular} &        & Classification                                                   & \multicolumn{1}{l}{10K}                                           & A patent dataset focusing on the crop industry domain. Available at: \url{https://github.com/GT4SD/domain-adaptive-patent-classifier}                                                                                                \\
DeepPatent \citep{kucer2022deeppatent}                                                                          & USPTO  & \begin{tabular}[c]{@{}l@{}}Patent\\ retrieval\end{tabular}       & 45K                                                               & Image-based patent retrieval benchmark. Available at: \url{https://github.com/GoFigure-LANL/DeepPatent-dataset}                                                                                                                      \\
PatentMatch \citep{risch2020patentmatch}                                                                        & EPO    & \begin{tabular}[c]{@{}l@{}}Patent\\ retrieval\end{tabular}       & 6.3M                                                              & Pairs of claims and text passages from cited patents. Available at: \url{https://hpi.de/naumann/s/patentmatch}                                                                                                                       \\
\citet{helmers2019automating}                                                                                   &        & \begin{tabular}[c]{@{}l@{}}Patent\\ retrieval\end{tabular}       & \multicolumn{1}{l}{2.5M}                                          & Patent pairs with labels of cited/random. Available at: \url{https://github.com/helmersl/patent_similarity_search}                                                                                                                \\

TFH \citep{chen2020deep}                                                                                        & USPTO  & \begin{tabular}[c]{@{}l@{}}Information\\ extraction\end{tabular} & \multicolumn{1}{l}{40K}                                           & Labeled patent dataset for entities and the semantic relations extraction. Available at: \url{https://github.com/awesome-patent-mining/TFH_Annotated_Dataset}                                                                        \\
\citet{arts2021natural}                                                                                         &        & \begin{tabular}[c]{@{}l@{}}Novelty\\ prediction\end{tabular}     & \multicolumn{1}{l}{}                                              & Patents with high and low novelty. Available at: \url{https://zenodo.org/records/3515985}                                                                                                                                            \\
\citet{jiang2023deep}                                                                                           & USPTO  & \begin{tabular}[c]{@{}l@{}}Acceptance\\ prediction\end{tabular}  & \multicolumn{1}{l}{5.5M}                                          & Granted and rejected patent applications. Available at: \url{https://github.com/Catchy1997/DLPAP}                                                                                                                                    \\
\citet{wu2023multi}                                                                                             & USPTO  & \begin{tabular}[c]{@{}l@{}}Litigation\\ prediction\end{tabular}  & 117K                                                              & Triples of a plaintiff, defendant, and a patent. Available at: \url{https://github.com/USTCwuhan/MANTF-master}                                                                                                                       \\
BigPatent \citep{sharma2019bigpatent}                                                                           & USPTO  & Summarization                                                    & 1.3M                                                              & Patent documents with human-written abstractive summaries. Available at: \url{https://evasharma.github.io/bigpatent/}                                                                                                                \\
\citet{casola2023creating}                                                                                      &        & Simplification                                                   & \multicolumn{1}{l}{0.2M}                                          & Silver standard dataset for patent simplification. Available at: \url{https://github.com/slvcsl/patentSilverStandard}                                                                                                                \\
GED \citep{wirth2023building}                                                                                   & EPO    & Translation                                                      & \multicolumn{1}{l}{19K}                                           & Patent language pairs French–English, and German–English. Available at: \url{https://huggingface.co/datasets/mwirth-epo/epo-nmt-datasets}                                                                                            \\
EuroPat \citep{heafield2022europat}                                                                             & EPO    & Translation                                                      & \multicolumn{1}{l}{51M}                                          & 6 official European languages paired with English. Available at: \url{https://opus.nlpl.eu/EuroPat.php}                                                                                                                              \\
\citet{lee2020patentgenerate}                                                                                   & USPTO  & Patent writing                                                   & \multicolumn{1}{l}{0.6M}                                          & Patent claims of the granted U.S. utility patents in 2013. Data information at: \url{https://doi.org/10.1016/j.wpi.2020.101983}                                                                                                      \\
HUPD \citep{suzgun2022harvard}                                                                                  & USPTO  & Multi-purpose                                                    & 4.5M                                                              & A large-scale patent corpus containing 34 data fields. Available at: \url{https://patentdataset.org/}                                                                                                                                \\ \bottomrule
\end{tabular}}
\caption{Curated patent datasets. }
\label{patentdata}
\end{table*}

\textbf{Datasets. } Patent offices provide large-scale raw data in the patent domain. Developers and researchers rely on well-curated datasets for development and research. We summarize representative publicly available curated patent datasets in Table~\ref{patentdata}. We aim to reduce the time and effort for data searching in the patent domain by presenting these ready-to-use datasets.

The number of curated datasets for patent classification and patent retrieval is typically larger than others because the data collection process is simple. Each granted patent is assigned classification codes and contains referenced patents. Hence, researchers can formulate the datasets by collecting and filtering patents without further labeling. For patent novelty quantification and prediction, \citet{arts2021natural} considered patents connected to awards such as the Nobel Prize as novel because they radically impacted technological progress and patenting. In contrast, patents were considered lacking novelty if they were granted by the United States Patent and Trademark Office but simultaneously rejected by both the European Patent Office and Japan Patent Office. However, this collection may deviate from the formal definition of novelty introduced in Section~\ref{novel_definition}. For patent simplification, there is only a silver standard dataset \citep{casola2023creating}. In text generation tasks, texts written by humans are usually considered the gold standard. Since patent simplification requires extensive expertise and effort, it is expensive and time-consuming to obtain a gold standard for patent simplification. Thus, \citet{casola2023creating} adopted automated tools to generate simpler texts for patents and named it the silver standard. 

Notably, the Harvard USPTO Patent Dataset (HUPD) \citep{suzgun2022harvard} is a recently presented large-scale multi-purpose dataset. It contains more than 4.5 million patent documents with 34 data fields, providing opportunities for various tasks. The corresponding paper demonstrates four types of usages of this dataset, including granting prediction, subject classification, language modeling, and summarization.

\noindent\textbf{Shared Tasks. } Some organizations proposed shared tasks and workshops in the patent domain to facilitate related research. Every participant worked on the same task with the same dataset to enable comparisons between different approaches.

The intellectual property arm of the Conference and Labs of the Evaluation Forum (CLEF-IP)\footnote{CLEF: \url{https://www.clef-initiative.eu/}} focuses on evaluation tasks related to intellectual property, particularly in patent retrieval and analysis. Each task usually contains a curated patent dataset for desired aims, including various information such as text, images, and metadata \citep{piroi2017evaluating}. 
    
The Japanese National Institute of Informatics Testbeds and Community for Information access Research (NTCIR)\footnote{NTCIR: \url{https://research.nii.ac.jp/ntcir/index-en.html}} provides datasets and organizes shared tasks to facilitate research in information retrieval, natural language processing, and related areas. Patent-related tasks at the NTCIR range from patent retrieval and classification to text mining and machine translation \citep{utiyama2007japanese, lupu2017patent}. 
    
TREC-CHEM\footnote{TREC-CHEM: \url{https://trec.nist.gov/data/chem-ir.html}} is a part of the Text REtrieval Conference (TREC) series, specializing in chemical information retrieval. It contains patent datasets that are rich in chemical information. For example, the dataset from TREC-CHEM 2009 contains 2.6 million patent files registered at the European Patent Office, United States Patent and Trademark Office, and World Intellectual Property Organization \citep{lupu2009trec}.

\section{Patent Analysis Tasks}
\label{analysistask}

\begin{table*}[!ht]
\footnotesize
\centering

\begin{tabular}{lll}
\toprule
Structure & Label      & \multicolumn{1}{c}{Description}   \\ 
\midrule
Section   & F          & Mechanical engineering ; Lighting; Heating; Weapons; Blasting  \\
Class     & F02        & Combustion engines; hot-gas or combustion-product engine plants \\
Sub-class & F02D       & Controlling combustion engines         \\
Group     & F02D 41    & Electrical control of supply of combustible mixture of its constituents \\
Sub-group & F02D 41/02 & Circuit arrangements for generating control signals \\ 
\bottomrule
\end{tabular}
\caption{Example of International Patent Classification (IPC) scheme}
\label{ipc_scheme}
\end{table*}

\begin{figure*}[!htbp]
    \centering
    \includegraphics[width=\textwidth]{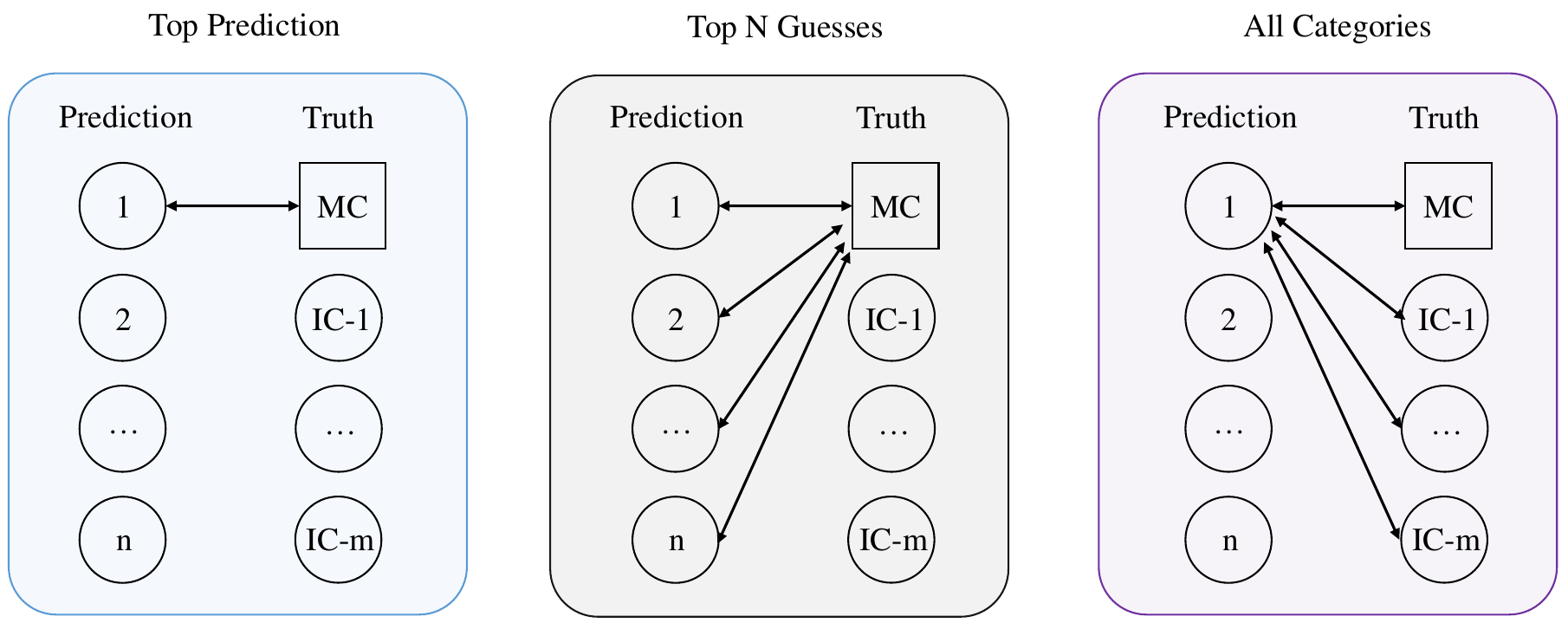}
    \caption[]{Illustration of three evaluation methods for patent classification, where 1, 2, ..., n are the top-n predictions, MC stands for main class, and IC is incidental class \citep{fall2003automated}. }
    \label{fig:class_eval}
\end{figure*}

Patent analysis tasks focus on understanding and usage of patents. We divide patent analysis tasks into four main types, subject classification, information retrieval, quality assessment, and technology insights. Patent subject classification (Section~\ref{classification}) is one of the most widely studied topics in the patent domain, where the categories of patents are predicted based on their content. Information retrieval tasks consist of two sub-tasks, specifically patent retrieval (Section~\ref{patentretrieval}) and information extraction (Section~\ref{informationextraction}). While patent retrieval aims at retrieving target documents from databases, information extraction focuses on extracting desired information from patent texts for further applications. Quality assessment refers to evaluating the quality of patents, which includes novelty prediction (Section~\ref{noveltyprediction}), granting prediction (Section~\ref{acceptanceprediction}), litigation prediction (Section~\ref{litigationprediction}), and patent valuation (Section~\ref{valuation}). As novelty is essential for patents, early prediction of novelty and auxiliary methods for novelty assessments could ensure patent quality before filing and improve efficiency. Granting prediction forecasts whether the patent office will grant a patent application. The process involves further aspects beyond novelty and inventiveness, such as formal requirements of the language, figures, and documents. A low-quality patent is likely to be rejected by the examiner. Litigation prediction measures the odds that the file may at some point become the subject of litigation. For example, patents with unclear or ambiguous claims or very scarce descriptions tend to be more likely to cause litigation cases. Patent valuation refers to measuring the value of the patents, which is a reflection of patent quality and scope. Technology insights are the usage of patents, consisting of technology forecasting (Section~\ref{technologyforecast}) and innovation recommendation (Section~\ref{innovationreccomendation}). Since patents contain extensive emerging technology information, researchers can analyze patents to predict future technological development trends or suggest new ideas for technological innovation. 

\subsection{Automated Subject Classification}
\label{classification}

\subsubsection{Task Definition of Subject Classification} 
\label{class_task}
The automated subject classification task is a multi-label classification task. The aim is to predict patents' specific categories or classes based on patent content, including title, abstract, and claims. Given a sequence of inputs $x=[w_1, w_2, \ldots, w_n]$, the objective is to predict the label $y \in \{y_1, y_2, \ldots, y_m\}$. Since patents may belong to multiple classes, sometimes more than one label needs predicting. This classification is crucial for organizing patent databases, facilitating patent searches, and assisting patent examiners in evaluating the novelty. 

\subsubsection{Classification Scheme} 
\label{class_scheme}
Two of the most popular classification schemes are the International Patent Classification  (IPC) and Cooperative Patent Classification (CPC) systems. These IPC/CPC codes are hierarchical and divided into sections, classes, sub-classes, main groups, and sub-groups. For example, we list the breakdown of the F02D 41/02 label using the IPC scheme in Table \ref{ipc_scheme}.

\begin{figure*}[!htbp]
    \centering
    \includegraphics[width=.8\textwidth]{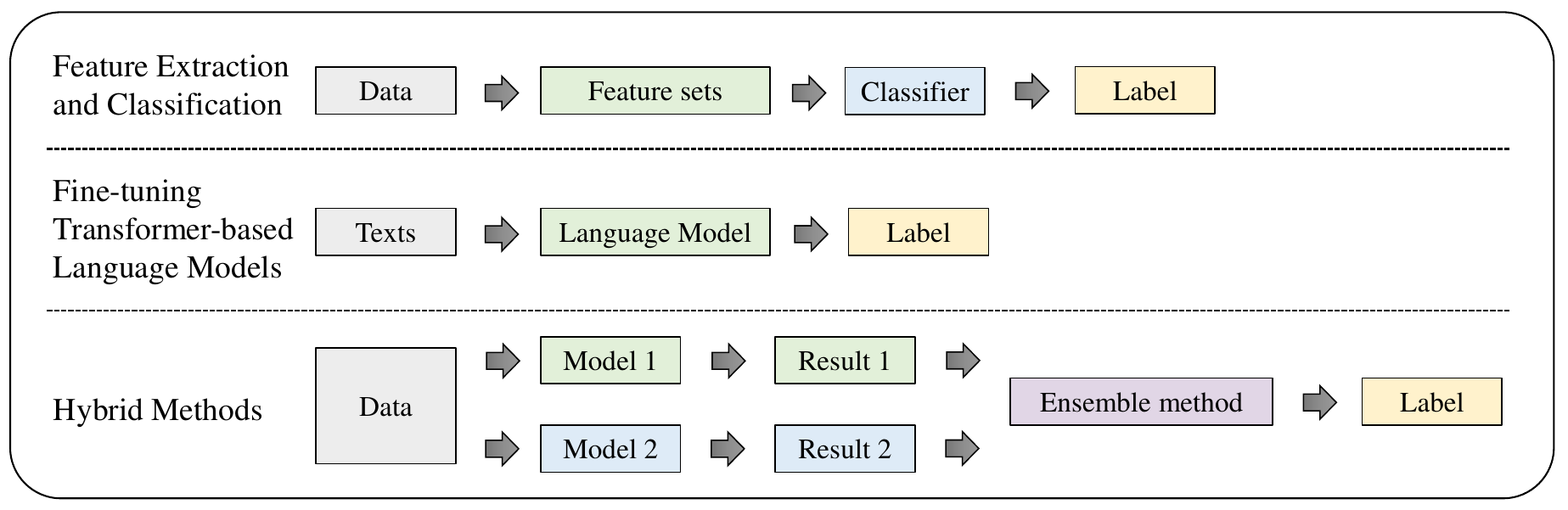}
    \caption[]{Three mainstream methods for patent classification }
    \label{fig:class_method}
\end{figure*}

\subsubsection{Evaluation}
\label{class_eval}
Most research uses one or more evaluation measures originated from \citet{fall2003automated}. As  Fig.~\ref{fig:class_eval} illustrates, there are three different evaluation methods, namely top prediction, top N guesses, and all categories. Top prediction only checks whether the top-1 prediction is the same as the main class. In top N guesses, the result is successful if one of the top-n predictions matches the main class, which is more flexible compared to the top-1 prediction. On the other hand, the all-categories method checks whether the top-1 prediction is included in the set of the main class and all incidental classes.

\subsubsection{Methodologies for Automated Subject Classification} 
\label{class_method}
We summarize the mainstream techniques for patent classification and categorize them into three types, including feature extraction and classification, fine-tuning transformer-based language models, and hybrid methods (Fig.\ \ref{fig:class_method}).

\textbf{Feature Extraction and Classifier. } Researchers extract various features from patent documents and adopted a classifier for prediction based on the features \citep{shalaby2018lstm, hu2018hierarchical, abdelgawad2019optimizing, zhu2020patent}. 

Most of the research used text content for prediction. The commonly used text representation is word embedding. \citet{roudsari2021comparison} compared five different text embedding approaches, including bag-of-words, GloVe, Skip-gram, FastText, and GPT-2. Bag-of-words is based on word frequency and ignores semantic information. GloVe, Skip-gram, and FastText are deep learning methods that capture word meaning but not word contexts. In contrast, GPT-2 is based on transformer architecture and is capable of capturing complex contextual information. The results demonstrated that GPT-2 performed the best with a precision of 80.52\%, indicating that transformer-based embedding approaches may perform better than other traditional or deep-learning embeddings. In addition, word embeddings are typically pre-trained on heterogeneous corpora, such as Wikipedia, making them lack domain awareness. Therefore, \citet{risch2019domain} trained a domain-specific word embedding for patents and incorporated recurrent neural networks for patent classification. This method improved the precision by 4\% compared to normal word embedding, demonstrating the effectiveness of domain adaptation. Apart from word embeddings, researchers also adopted sentence embedding for feature extraction \citep{bekamiri2021patentsberta}. Other research calculated the semantic similarity between patent embeddings obtained from Sentence-BERT \citep{reimers2019sentence} and used the k-nearest neighbors (KNN) method for classification with a precision of 74\%. Although the performance was not strong enough, the use of similarity and KNN provided a different view of the classification approach. 

\begin{table*}[!htbp]
\resizebox{\textwidth}{!}{

\centering

\begin{tabular}{lllllllc}
\toprule
Paper                                & \begin{tabular}[c]{@{}l@{}}Data \\ source\end{tabular} & \begin{tabular}[c]{@{}l@{}}Dataset\\ size\end{tabular} & Parts used                                                                                            & \begin{tabular}[c]{@{}l@{}}Number of \\ classes\end{tabular} & Methods                                                                                  & Metric & \begin{tabular}[c]{@{}c@{}}Results\\ (Top prediction)\end{tabular} \\ \midrule
\multicolumn{8}{l}{Feature Extraction and Classification}                                                                                                                                                                                                                                                                                                                                                                                                                                              \\
\citep{risch2019domain}              & USPTO-2M                                               & 1.95, 0.05                                             & Title, abstract                                                                                       & 637                                                          & \begin{tabular}[c]{@{}l@{}}Domain-specific\\ word embedding + RNN\end{tabular}           & P     & 53.00                                                              \\
\citep{li2018deeppatent}             & USPTO-2M                                               & 1.95, 0.05                                             & Title, abstract                                                                                       & 637                                                          & Skip-gram + CNN                                                                          & P      & 73.88                                                              \\
\citep{roudsari2021comparison}       & USPTO                                                  & 0.24, 0.04                                             & Title, abstract                                                                                       & 89                                                           & GPT-2 + CNN                                                                              & P      & 80.52                                                              \\
\citep{bekamiri2021patentsberta}     & USPTO                                                  & 1.37, 0.12                                             & Title, abstract, claims                                                                               & 663                                                          & SBERT + KNN                                                                              & P      & 74.00                                                              \\
\citep{jiang2021deriving}            & USPTO                                                  & 0.35, 0.05                                             & Image                                                                                                 & 8                                                            & CNN                                                                                      & A      & 54.32                                                              \\ \midrule
\multicolumn{8}{l}{Fine-tuning Transformer-based Language Models}                                                                                                                                                                                                                                                                                                                                                                                                                                      \\
\citep{lee2020patent}                & USPTO-2M                                               & 1.95, 0.05                                             & Title, abstract                                                                                       & 637                                                          & BERT                                                                                     & P      & 81.75                                                              \\
\citep{haghighian2022patentnet}      & USPTO-2M                                               & 1.95, 0.05                                             & Title, abstract                                                                                       & 544                                                          & XLNET                                                                                    & P      & 82.72                                                              \\
\citep{christofidellis2023automated} & USPTO                                                  & 0.24, 0.04                                             & Title, abstract                                                                                       & 89                                                           & \begin{tabular}[c]{@{}l@{}}SciBERT + Domain-adaptive \\ training + adapters\end{tabular} & P      & 84.53                                                              \\
\citep{risch2020hierarchical}        & USPTO-2M                                               & 1.95, 0.05                                             & Title, abstract                                                                                       & 632                                                          & Transformer                                                                              & A      & 56.70                                                              \\ \midrule
\multicolumn{8}{l}{Hybrid Methods}                                                                                                                                                                                                                                                                                                                                                                                                                                                                     \\
\citep{jiang2022deep}                & USPTO                                                  & 0.72, 0.08                                             & Title, abstract, image                                                                                & 622                                                          & Multimodal                                                                               & A      & 65.50                                                              \\
\citep{zhang2022reliable}            & Chinese                                                & 0.61, 0.15                                             & Title, abstract                                                                                       & 8                                                            & Multi-view learning                                                                       & A      & 85.21                                                              \\
\citep{kamateri2023ensemble}         & CLEFIP-0.54M                                           & 0.49, 0.05                                             & \begin{tabular}[c]{@{}l@{}}Title, abstract, description,\\ claims, applicants, inventors\end{tabular} & 731                                                          & Ensemble methods                                                                         & A      & 70.71                                                              \\ \bottomrule
\end{tabular}}
\caption{List of representative papers for patent classification. Note: The first number in dataset size is the sum of training and validation samples, and the second number is the size of the test set. The unit here is million. The metric P stands for precision, while A stands for accuracy. There are multiple results from each paper, and methods with the best performance are selected.}

\label{tab:class_method}
    
\end{table*}

In addition, some research studied variations of network architecture and optimization methods. \citet{shalaby2018lstm} improved original paragraph vectors \citep{le2014distributed} to represent patent documents. The authors used the inherent structure to derive a hierarchical description of the document, which was more suitable for capturing patent content. In addition, \citet{abdelgawad2019optimizing} analyzed hyper-parameter optimization methods for different neural networks, such as CNN, RNN, and BERT. The results illustrate that optimized networks could sometimes yield 6\% accuracy improvement. Similarly, \citet{zhu2020patent} used a symmetric hierarchical convolutional neural network and improved the F1 score by approximately 2\% on the Chinese short text patent classification task compared to a conventional convolutional neural network. 

Alternatively, images can also serve for automated patent classification, which is introduced in Appendix \ref{a:classification}.

\textbf{Fine-Tuning Transformer-Based Language Model. } Fine-tuning tailors the pre-trained model to the patent classification task. Lee et al.\ fine-tuned the BERT model for the patent classification task and achieved 81.75\% precision \citep{lee2020patent}, which was more than 7\% higher than traditional machine learning under the same setting, using word embedding and classifiers \citep{li2018deeppatent}. Moreover, \citet{haghighian2022patentnet} compared multiple transformer-based models, including BERT, XLNet, RoBERTa, and ELECTRA, indicating that XLNet performed the best regarding precision, recall, and F1 score. Furthermore, research found that incorporating further training approaches can optimize the fine-tuning process. For example, \citet{christofidellis2023automated} integrated domain-adaptive pre-training and used adapters during fine-tuning, which improved the final classification results by about 2\% of F1 score. Transformer-based language models have shown better effectiveness than traditional text embedding and become the mainstream method for text-based problems. 

Another study addressed this hierarchical classification task as a sequence generation task. \citet{risch2020hierarchical} implemented the transformer-based models to follow the sequence-to-sequence paradigm, where the input was patent texts and the output was the class, such as F02D 41/02. However, the highest accuracy tested on their datasets was only 56.7\%, which demonstrated that the sequence-to-sequence paradigm for patent classification was mediocre.

\textbf{Hybrid Methods. } Hybrid methods refer to combining different approaches to make predictions. TechDoc \citep{jiang2022deep} is a multimodal deep learning architecture, which synthesizes convolutional, recurrent, and graph neural networks through an integrated training process. Convolutional and recurrent neural networks served to process image and text information respectively, while graph neural networks served for the relational information among documents. This multimodal approach was able to reach greater classification accuracy than previous unimodal methods. The advantage of multimodal models is to leverage different modalities for a comprehensive prediction. 

Additionally, \citet{zhang2022reliable} used a multi-view learning method \citep{zhao2017multi} for patent classification. Multi-view learning integrates and learns from multiple distinct feature sets or views of the same data, aiming to improve model performance by leveraging the complementary information available in different views. In a general multi-view learning pipeline, developers specify a model in each view. They aggregate and train all models collaboratively based on multi-view learning algorithms.  \citet{zhang2022reliable} used the two views of patent titles and patent abstracts and tested multi-view learning on a Chinese patent dataset, demonstrating its effectiveness and reliability. 

Moreover, a recent study investigated ensemble models that can combine multiple classifiers for classification \citep{kamateri2023ensemble}.  While multi-view learning exploits diverse information from different data sources for a more comprehensive understanding, ensemble methods focus on combining predictions from multiple models to improve prediction accuracy and robustness by reducing the models' variance and bias, which may all use the same data view. The authors experimented with different ensemble methods and finally achieved 70.71\% accuracy, which was higher than using only one classifier.

For a synopsis, we list some representative papers in Table \ref{tab:class_method}, including data sources, dataset size, parts used for training, number of classes, methods, and results. The table shows that researchers use different datasets and metrics to test their models, which complicates the comparison of various methods. Hence, we call for standard benchmarks for patent classification to facilitate the development.  In addition, LLMs demonstrate more promising results compared to traditional ML models. However, most research is still based on outdated models, such as GPT-2. The application of recent large-sized LLMs to this task would enhance the effectiveness. Additionally, most research has focused on short texts for patent classification, such as abstracts and claims. Nonetheless, titles and abstracts of patents are generic and do not disclose much relevant information. We recommend future research focus more on detailed patent descriptions that contain detailed and specific information about an invention. Furthermore, domain-adaptive methods that adapt standard LLMs to the patent domain are worth investigating to optimize the performance.

\subsection{Patent Retrieval}
\label{patentretrieval}
There are three types of retrieval tasks, prior-art search, patent landscaping, and freedom to operate search. Since previous research mainly focused on prior-art search, we introduce this task first in Section~\ref{pas_task} with its corresponding methods in Section~\ref{pas_method}, followed by patent landscaping in Section~\ref{landscape} and freedom to operate search in Section \ref{s:free}.

\subsubsection{Task Definition of Prior-Art Search}
\label{pas_task}
Prior-art search refers to, given a target patent document $X$, automatically retrieving $K$ documents that are the most relevant to $X$ from a patent database.\footnote{While most research retrieved patent documents as outputs, prior art can also be other types of published (oral or visual) information. } This process is crucial for patent examiners to assess the patentability of a new patent application. Prior-art search is not trivial, due to the intricate patent language and the different terms used in various patent descriptions. In principle, documents from distant fields, including both patent literature and other sources, can compromise the novelty of a new application if they disclose the same combination of patent features. However, these different fields often use distinct nomenclature across all word classes, including nouns, verbs, and adjectives. Additionally, many patents and patent applications create their own terms, which can significantly differ from everyday language and even from the terminology used in the technical field. This creation of terms is not always intentional. Attorneys and patent professionals, who may not be deeply familiar with a specific field, might need precise terms for their descriptions and claims and thus decide to invent names spontaneously. These self-defined terms are often highly artificial and may not appear in any other relevant documents. Such terms do not need to be listed in any dictionary.

\begin{figure*}[!htbp]
    \centering
    \includegraphics[width=\textwidth]{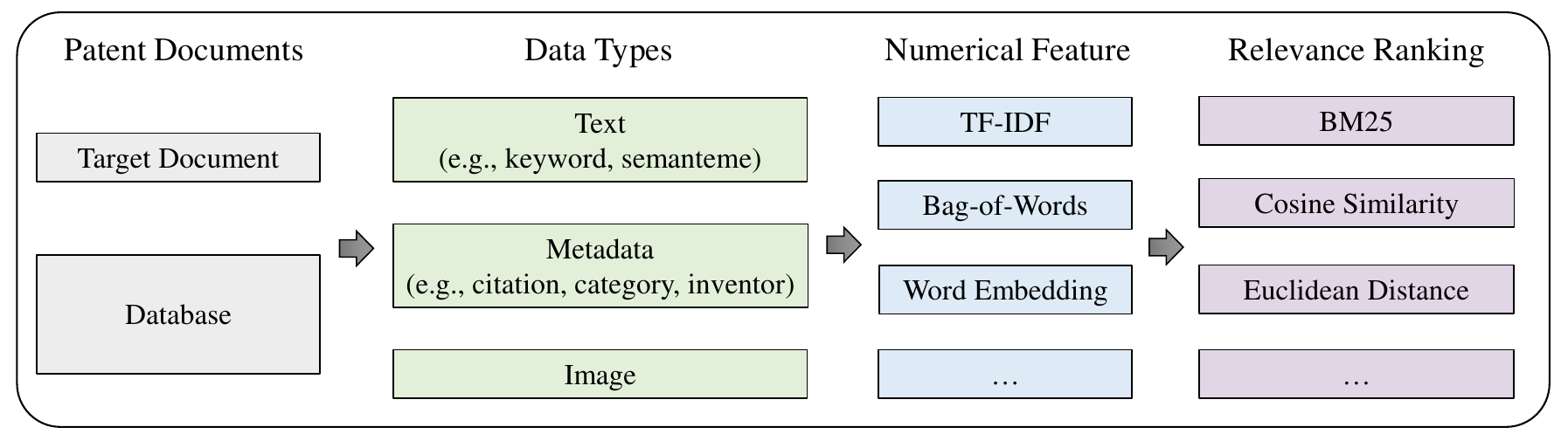}
    \caption[]{Illustration of the patent retrieval process}
    \label{fig:retrival_method}
\end{figure*}

\subsubsection{Methodologies for Prior-Art Search} 
\label{pas_method}
Researchers have intensively invested in patent retrieval tasks and achieved some encouraging accomplishments. We summarize the general retrieving process in Fig.~\ref{fig:retrival_method}. Data types for further pre-processing include text, metadata, and images. Multiple methods can transform patent data into numerical features, such as the statistical method term frequency-inverse document frequency (TF-IDF), which is widely known from text-based search systems, and deep-learning-based word embedding. Relevance ranking algorithms to retrieve the most relevant documents include best-matching BM25, which is a bag-of-word-type method, and cosine similarity derived from the inner vector product. 

We focus on text-based patent retrieval in this section and introduce other methods in Appendix~\ref{a:retrieval}. Traditional methods are keyword-based searching and statistical approaches that can rank document relevance, such as the BM25 algorithm \citep{robertson2009probabilistic}. Keyword-based methods refer to a search for exact matches in the target corpus according to the input query. A previous survey summarized all keyword-based methods into three categories, namely query expansion, query reduction, and hybrid methods \citep{shalaby2019patent}. On the other hand, statistical methods exploit a document's statistics, such as the frequency of specific terms, to calculate a relevance score based on the inputs. For example, the BM25 algorithm calculates the sore using the occurrence of the query terms in each document. These methods are straightforward but the significant semantic and context information has not been explored. Furthermore, each patent document may use its own nomenclature, which may be defined in the description counter-intuitively to daily use.

Researchers proposed two types of improvement to enhance the patent retrieving performance, from keyword-based to full-text-based and from statistical methods to deep-learning methods. Errors occur inherently in keyword-based methods because different keywords can represent the same technical concepts across various disciplines. Hence, \cite{helmers2019automating} adopted entire patent texts for similarity comparison and evaluated various feature extraction methods, such as bag-of-words, Word2Vec, and Doc2Vec. The results demonstrated that full-text similarity search could bring better retrieving quality. Moreover, a large body of research studied different deep learning-based embeddings that transfer texts to numerical values for similarity calculation \citep{sarica2019engineering, hain2022text, hofstatter2019enriching, deerwester1990indexing, althammer2021cross, trappey2021intelligent, vowinckel2023searchformer}. Since word embeddings cannot capture contextual information at a higher level, \cite{hofstatter2019enriching} improved the Word2Vec model by incorporating global context, yielding up to 5\% increase in mean average precision (MAP). At the paragraph level, \cite{althammer2021cross} evaluated the BERT-PLI (paragraph-level interactions), which is specifically designed for legal case retrieval  \citep{shao2020bert}, in both patent retrieval and cross-domain retrieval tasks. However, the authors showed that the performance did not surpass the BM25 baseline and indicated that BERT-PLI was not beneficial for patent document retrieval. In another research, \cite{trappey2021intelligent} trained the Doc2Vec model, an embedding that can capture document-level semantic information, based on patent texts for patent recommendation. Patent recommendation aims to retrieve target patent documents from the database. The results showed that Doc2Vec led to more than 10\% improvement compared to bag-of-words methods or word embeddings. This research suggests that document-level embeddings are more promising for document retrieval because they can effectively capture context information. 

LLMs have demonstrated effectiveness in text retrieval tasks \citep{ma2023fine}. Thus, using LLMs for patent prior-art search is a promising research direction. In addition, studies have shown that integrating retrieval into LLMs can improve factual accuracy \citep{nakano2021webgpt}, downstream task performance \citep{izacard2023atlas}, and in-context learning capabilities \citep{huang2023raven}. These retrieval-augmented LLMs are well-established for handling question-answering tasks \citep{xu2024retrieval}. The application of retrieval-augmented generation in the patent domain is another interesting research direction.

\subsubsection{Patent Landscaping}
\label{landscape}
Patent landscaping aims to retrieve patent documents related to a particular topic. Landscaping might have a larger overall strategic value for companies and be more related to the machine learning topic. However, patent landscaping is much less investigated compared to prior-art search.

It is straightforward to relate landscaping to prior-art search in two ways. Firstly, we can consider the target topic as a keyword for patent retrieval and use keyword-based methods to retrieve documents from the database. Alternatively, we can find seed patents to represent the topic and retrieve documents that are related to the seed patents as the result \citep{abood2018automated}. 

Furthermore, researchers developed classification models for patent landscaping, which classify whether a patent belongs to a given topic \citep{choi2022deep, pujari2022three}. \cite{choi2022deep} concatenated text embedding of abstracts and graph embedding of subject categories for patent representations. Subsequently, the authors added a simple output layer to conduct the necessary binary classification task. To stimulate the research on patent-landscaping-oriented classification, \cite{pujari2022three} released three labeled datasets with qualitative statistics. 

\subsubsection{Freedom-to-Operate Search}
\label{s:free}
The freedom-to-operate (FTO) search, also known as the clearance search, is a specific type of patent-related research. The aim is to determine if a particular technology or product would be covered by any active intellectual property rights of another party. This search is critical for companies before launching a new product or service in the market and may be requested by investors as part of a due-diligence process.

This task shares similarities with a prior-art search but has an important difference: The search for prior art takes the technology in question and checks if any prior record (not limited to patent documents or active patents) alone (novelty) or in combination (inventiveness/obviousness) contains all the features of the technology. Thus, the search analyzes if the technology is entirely part of the prior art. The documents or other records of the prior art anticipating the technology may also contain more features in their claims.  In contrast, the search for freedom-to-operate checks if there is any active patent (or application still in examination), where an independent claim has fewer features (constituting a more general invention) than the technology in question requires. Thus, individual claims of relevant prior-art documents are practically in their entirety included in the technology. Accordingly, while a technology might be patentable due to its novelty and inventiveness, it could still be covered by an earlier patent or pending application if it incorporates the features of the prior art alongside some additional nonobvious features. Therefore, the new invention would be classified as a more specific dependent invention.

Following the task definition, a freedom-to-operate search needs to analyze the claims of potentially relevant patents (applications) in detail and break down the technology in question. Hence, the automated retrieval of targeted patent claims is the core of this task. Few studies have investigated this type of retrieval task. \citet{freunek2021bert} trained BERT for the freedom-to-operate search process. They cut patent descriptions into pieces and use BERT to retrieve relevant claims from a constructed dataset. Their report demonstrated that BERT was able to identify relevant claims in small-scale experiments. As this task is important but widely neglected by the community, we introduce it here and suggest it for future research.

\subsection{Information Extraction}
\label{informationextraction}

\subsubsection{Task Definition of Information Extraction}
\label{IE_task}
\begin{figure*}[!htbp]
    \centering
    \includegraphics[width=\textwidth]{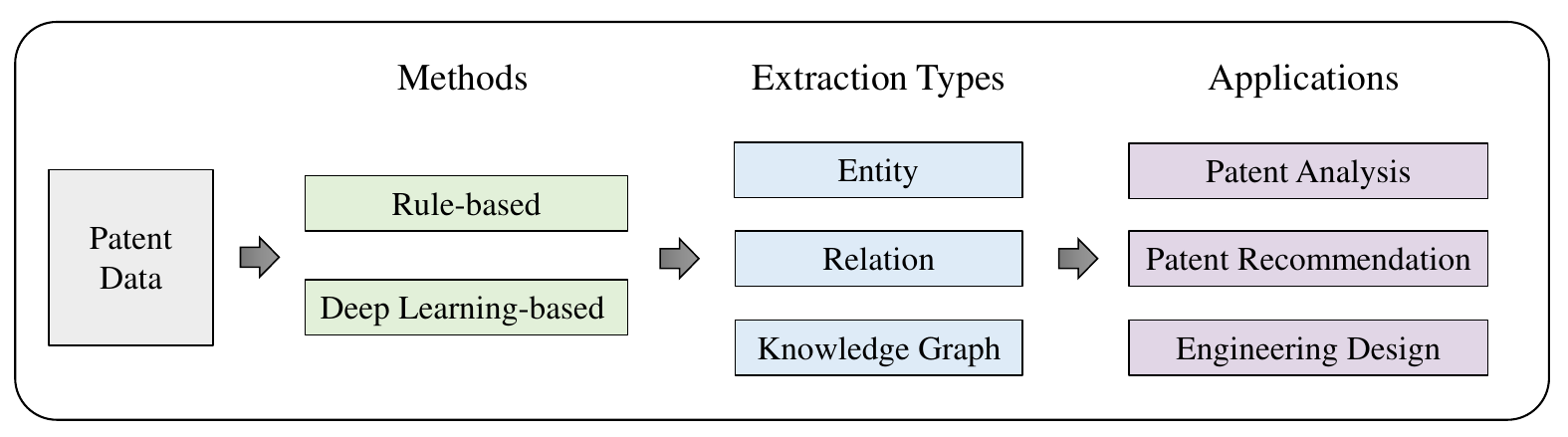}
    \caption[]{Demonstration of information extraction process in patent domain }
    \label{fig:ie_method}
\end{figure*}

The process of extracting specific information from a text corpus is called information extraction. The goal of information extraction is to transform textual data into a more structured format that can be easily processed for various applications, such as data analysis. Hence, researchers usually use information extraction as a support task for patent analysis. Fig.~\ref{fig:ie_method} demonstrates how information extraction is applied in the patent domain. Rule-based and deep-learning-based methods are two main streams of information extraction. The extracted information can be entities, relations, or knowledge graphs, which are constructed based on entities and relations. This information can serve for further tasks, such as patent analysis and patent recommendations for companies. 

\subsubsection{Methodologies for Information Extraction}
\label{IE_method}
\textbf{Rule-Based Methods. } Traditional extraction methods are rule-based. Researchers manually pre-define a set of rules and extract desired information based on the rules. \citet{chiarello2019approaches} designed a rule-based system to extract affordances from patents. For example, one of the rules was ``The term user followed by can and adverbs, such as readily efficiently, quickly and easily''. The authors used the extracted results to evaluate the quality of engineering design. Another study devised rules according to the syntactic and lexical features of claims to extract facts \citep{siddharth2022engineering}. The authors integrated and aggregated these facts to obtain an engineering knowledge graph, which could support inference and reasoning in various engineering tasks. 

Well-defined rules can lead to precise extraction so that this process is transparent and interpretable. However, creating and maintaining rules can be time-consuming, difficult, and biased. Moreover, rule-based methods may struggle with the variability and complexity of natural language.

\textbf{Deep-Learning-Based Methods. } Deep learning methods require labeled datasets that include entities or relations for training. There are different network architectures for model training, such as long short-term memory (LSTM) \citep{chen2020deep} and transformers \citep{son2022ai, puccetti2023technology, giordano2022unveiling}. Notably, \citet{son2022ai} stated that most patent analysis research focused on claims and abstracts, neglecting description parts that contain essential technical information. The reason may be the notably larger length of the description, which requires appropriate models that can load such text length. Thus, the authors proposed a framework for information extraction through patent descriptions based on the T5 model. 

Deep-learning methods can handle a wide variety of language expressions and are easily scalable with more data and computational power. Moreover, deep learning can capture complex patterns and dependencies in language. Nonetheless, deep learning requires large amounts of high-quality annotated data and computation resources for training.

\subsubsection{Applications of Information Extraction}
\label{IE_application}

\textbf{Patent Analysis. } Patent analysis involves the analysis of patents with respect to multiple aspects, such as evaluating patent novelty or quality, and forecasting technology trends. For example, \citet{chiarello2019approaches} extracted sentences with a high likelihood of containing affordance from patents to evaluate the quality of engineering design, for example: ``The user can easily navigate a set of visual representations of the earlier views.'' \citet{puccetti2023technology} identified technologies mentioned in patents to anticipate trends for an accurate forecast and effective foresight.

\textbf{Patent Recommendation. } Patent recommendation refers to suggesting relevant patents to users based on their interests, research, or portfolio. \citet{deng2021knowledge} extracted the semantic information between keywords in the patent domain and constructed weighted graphs of companies and patents. The authors compared the distance based on weighted graphs to generate recommendations. \citet{chen2023interpretable} extracted connectivity and quality attributes for pairs of patents and companies. Based on knowledge graphs and deep neural networks, the authors designed an interpretable recommendation model that improved the mean average precision of best baselines by 8.6\%. 

\textbf{Engineering Design. }  Engineering design is a creative process that involves defining a problem, conceptualizing ideas, and implementing solutions. The goal is to develop functional, efficient, and innovative solutions to meet specific requirements. Designers can gain insight by analyzing problems and principal solutions extracted from patent documents. \citet{giordano2022unveiling} adopted transformer-based models to extract technical problems, solutions, and advantageous effects from patent documents and achieved an F1 score of 90\%. The extracted information helps reveal valuable information hidden in patent documents and generate novel engineering ideas. Similarly, \citet{jiang2023extraction} used pre-trained models to identify the motivation, specification, and structure of inventions with the accuracy of 63\%, 56\%, 44\% respectively compared to expert analysis. From design intent to specific solutions, designers can review patents from a systematic perspective to gain better design insights.  

The recent LLMs have shown outstanding capabilities in information extraction, such as in complex scientific texts \citep{dunn2022structured} and medical domain \citep{pmlr-v225-goel23a}. Therefore, we anticipate the application of LLMs in the patent field to improve the quality of information extraction.

\subsection{Novelty and Inventiveness Prediction}
\label{noveltyprediction}

\subsubsection{Definition of Patent Feature, Novelty, and Inventiveness}
\label{novel_definition}
Novelty and inventiveness have a clear legal definition in most jurisdictions,\footnote{See Articles 54 and 56 of the European Patent Convention (EPC), Sections 2 and 3 of the UK Consolidated Patent Act, or §102 and §103 of the United States patent law codified in Title 35 Code of Laws (USC) as examples.} which may strongly deviate from common associations \citep{epo2020, uspto2022}. An invention is conceived as a combination of \textit{features}. It is novel if there is no older document or other form of disclosure\footnote{They collectively refer to the prior art or state of the art.} that alone includes and/or describes all essential \textit{features} of the invention.
Moreover, an invention may be considered novel for two more reasons. (1) The aggregation of prior disclosures does not reveal all its essential \textit{features}. (2) Experts in the field would not find it obvious to integrate any missing \textit{features}, for instance, according to their standard practice within the field.

Thus, the assessment of novelty critically relies on the concept of \textit{features}. The \textit{features} are the elements the invention needs to comprise to be the invention and are outlined in the patent claims. The independent claims list the essential \textit{features}. For example, \textit{features} can be physical elements and objects (typically nouns together with further specifiers), properties (often adjectives), or processing steps in a method. Importantly, \textit{features} in many jurisdictions cannot be negative, i.e., a missing property. Some offices allow the exclusion of a specific technology from the prior art through negative limitations if the description clearly states the absence of the feature. Later exclusion of features in the claims based on an absence in the description is not possible. 
The same feature can have very different names in different documents or even be denoted by a neologism well-defined in corresponding invention descriptions. The high variability of terminology between different documents is a major challenge for the examination and also for LLMs. Therefore, novelty prediction is a well-specified task by law and exhibits a high level of mathematical precision atypical of other language-related tasks. In contrast, inventiveness, determining whether the addition of certain \textit{features} to existing technology is obvious to an expert, can often be ambiguous.

\subsubsection{Task Definition of Novelty and Inventiveness Prediction}
\label{novel_task}
Novelty and inventiveness prediction is a binary classification task, aiming to determine whether a new patent is novel, given the existing patent database.\footnote{While researchers consider this task as a binary classification task, it may also be useful to assess the level of novelty of an invention. } Novelty is one of the essential requirements of patent applications and takes a vast of resources and time for human assessment. In addition, the process of reviewing patents is complex and detail-oriented. Even experienced examiners can overlook critical information or fail in judgment. Automated patent novelty evaluation systems can be used as an auxiliary tool for novelty examination. Therefore, this system is critical, because it can not only improve the quality of patents but also enhance the efficiency of patent examination. Since novelty prediction is substantially based on text analysis, the recent LLMs appear well-suited for this complex task.

\begin{figure*}[!htbp]
    \centering
    \includegraphics[width=.99\textwidth]{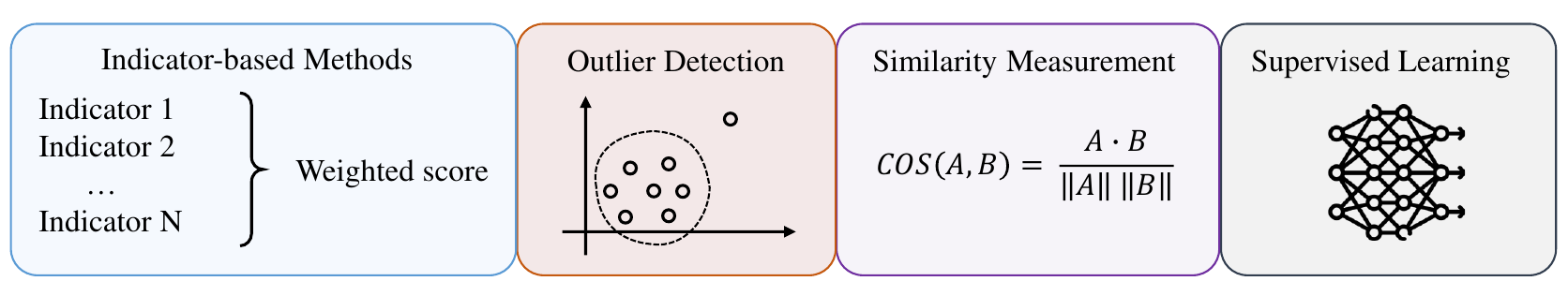}
    \caption[]{Methods for patent novelty prediction }
    \label{fig:novelty}
\end{figure*}

\subsubsection{Methodologies for Novelty and Inventiveness Prediction}
\label{novel_method}
\textbf{Indicator-Based Methods. } Indicator-based approaches rely on pre-defined indicators to measure patent novelty compared to prior art \citep{verhoeven2016measuring, plantec2021impact, sun2022measuring, wei2024extraction, schmitt2024modeling}. Researchers define indicators from various aspects, such as the citations, and assign each indicator a score or weight based on its importance to calculate novelty scores. 

For example, \citet{verhoeven2016measuring} proposed three dimensions to evaluate technological novelty, including novelty in recombination, novelty in technological origins, and novelty in scientific origins. They involved patent classification codes and citation information to analyze each indicator, showing that technological novelty in each dimension was interrelated but conveyed different information. Additionally, \citet{plantec2021impact} investigated technological originality, which was defined as the degree of divergence between underlying knowledge couplings embedded in the invention and the predominant design. The authors used proximity indicators of direct citations, co-citation, cosine similarity, co-occurrence, and co-classification, with normalization methods to balance different indicators. 

Notably, most research focused on evaluating technological novelty and originality, which was different from the formal definition of patent novelty. While patent novelty is assessed based on prior art and disclosures in a legal context, technological novelty and originality are analyzed in a technical context, focusing on the advancement and uniqueness of the technological contribution. In addition, indicator-based methods also pose some limitations. The selection and weighting of indicators can be subjective, and the indicator may not fully capture the nuanced aspects of inventions. 

\textbf{Outlier Detection. } Outlier detection is based on the assumption that novel inventions can be seen as outliers within the landscape of existing patents \citep{wang2019novelty, zanella2021understanding, jeon2022doc2vec}. Researchers used text embeddings to represent patents and applied outlier detection algorithms to identify patents that are different from the majority. 

Researchers used local outlier factor (LOF) for novelty outlier detection, which measured how isolated the object is with respect to the surrounding neighborhood \citep{breunig2000lof}. For example, \citet{zanella2021understanding} used Word2Vec \citep{mikolov2013efficient} to obtain text embedding based on patent titles and abstracts. \citet{wang2019novelty} extracted semantic information by using latent semantic analysis (LSA) \citep{deerwester1990indexing} based on patent titles, abstracts, and claims. These embeddings served as the input for LOF to measure the novelty of patents. However, this method deviates from the definition of patent novelty, because it does not measure outliers strictly by the features as suggested by the legal definition. Thus, these methods may correlate with the formal patent novelty but are not equivalent and therefore not necessarily useful for an automated examination process. 

Furthermore, \citet{jeon2022doc2vec} used Doc2Vec \citep{le2014distributed} to process patent claims as input for LOF. Since patent claims describe features of inventions, this method seems plausible according to the legal definition of novelty. Nonetheless, there are still some questions that are not explained. For instance, the difference between text embeddings may not stand for a feature difference of inventions. Therefore, using LLMs to explicitly extract and compare features between the target patent and the prior art is a more sensible approach for novelty prediction.  

Another limitation of LOF is that it sometimes flags patents that are unusual but not necessarily novel or original in a meaningful way. For example, a patent combining existing technologies in a statistically rare way may be classified as an outlier. However, such an outlier for statistical reasons without actual technical novelty does not represent a significant technological advancement.

\textbf{Similarity Measurements. } Researchers calculate similarities between the target patent and existing patents for novelty assessment \citep{siddharth2020toward, beaty2021automating, arts2021natural,shibayama2021measuring}. They used word embeddings based on the patent text, such as abstracts, and adopted various metrics to calculate the similarities, for example, cosine similarity and Euclidean distance. A patent with a low similarity score is considered more novel. 

Previous studies have investigated different text representation methods for similarity calculation. For example, \citet{arts2021natural} extracted keywords that related to the technical content of patents from titles, abstracts, and claims.  Each patent is represented in a 1,362,971 dimension vector, where each dimension was the frequency of a keyword. However, the authors considered patents with major impacts on technological progress as novel, which was different from the legal definition. 

Additionally, \citet{shibayama2021measuring} compared Word2Vec \citep{mikolov2013efficient} embeddings between the target patent and its cited patents based on abstracts, keywords, and titles.
Nonetheless, patent titles and abstracts are generic and do not disclose patent features, thereby failing to assess patent novelty in the legal sense.

\textbf{Supervised Learning. } Supervised learning refers to training models to classify whether a given patent is novel, based on target patent texts and the prior art \citep{chikkamath2020empirical, jang2023explainable}. 

\citet{chikkamath2020empirical} investigated a series of machine learning models for novelty detection. The authors conducted comprehensive empirical studies to evaluate the performance, including various text embeddings (e.g., Word2Vec \citep{mikolov2013efficient}, GloVe \citep{pennington2014glove}), different classifiers (e.g., support vector machine, naive Bayes), and multiple network architectures (e.g., LSTM \citep{hochreiter1997long}, GRU \citep{chung2014empirical}). The inputs are target patent claims and cited paragraph texts from prior art that are related to the target patent. This process makes sense because cited paragraphs possibly include features relevant to target patent claims, providing an implicit ground check for novelty. However, it is questionable whether the model detects novelty by comparing features rather than other aspects as deep learning models are typically black-box methods and lack explainability. 

To address the problem,  \citet{jang2023explainable} proposed an explainable model based on BERT \citep{devlin2018bert} to evaluate novelty. The authors aimed to follow the legal definition by comparing the claims of a target patent with its prior art. The model could output the novelty prediction result, along with claim sets with high relevance as an explanation, which achieved 79\% accuracy under the experimental settings. This paper presented a great idea, but BERT is nowadays outdated and outperformed by larger and more powerful LLMs.

\textbf{Suggestions for Future Work. } Based on the review of previous studies, we provide three suggestions for future research. (1) Researchers should differentiate between patent novelty and other similar terms, such as technology originality. Patent novelty focuses on the invention's features compared to the prior art in a legal context, whereas technological originality refers to the advancement and uniqueness of the technological contribution in the technical context. (2) Researchers should concentrate on specific patent content for assessing novelty. Using patent claims for prediction is a sensible approach as they include essential features for comparison. In contrast, studies that leverage titles and/or abstracts are not necessarily useful for an automated examination process, because they are usually generic and vain. (3) Researchers should explore the effectiveness of powerful LLMs, such as GPT-4, in this field. While LLMs have revolutionized the field of NLP, current studies on novelty prediction are still largely based on old-fashioned methods, such as word embeddings. Using LLM-based methods may significantly stimulate the automation of novelty prediction. 

\subsubsection{Patentability Assessment}
\label{novel_related}
Patentability refers to a set of criteria that an invention must meet to be eligible for a patent. The key criteria typically include novelty, inventiveness/non-obviousness, and utility. Novelty that has been discussed above means the invention has not been known or used in the prior art. Non-obviousness indicates that a patent should include a sufficient inventive step beyond what is already known, and should not be obvious to an expert in the field. Utility means the invention is useful and has some practical applications. Therefore, patentability assessment is a more comprehensive and challenging task compared to novelty prediction. 

While most works focused on novelty prediction, \citet{schmitt2023assessment} investigated patentability assessment by examining both novelty and non-obviousness following the legal definition. The authors used a mathematical-logical approach to decompose patent claims into features and formulate a feature combination. They compared the target patent feature combination with its prior art to evaluate novelty and non-obviousness based on the legal definition. Since the authors only processed one example patent for proof-of-concept, the effectiveness and efficiency of this method on large-scale patent applications are unknown. They pointed out another limitation that this method could not detect homonyms or synonyms. The recent LLMs with outstanding capability would undoubtedly contribute to this task both effectively and efficiently.

\subsection{Granting Prediction}
\label{acceptanceprediction}
\subsubsection{Task Definition of Granting Prediction}
Patent granting prediction refers to predicting whether a patent application will be granted or terminally rejected by examiners\footnote{A patent is typically terminally rejected if the inventors or their representatives cannot find any more features to establish novelty and inventiveness. Practically any good patent application is rejected in the first cycle. If a patent application is not rejected but granted based on the first draft of the claims, the patent attorneys---except for a few rare cases---should potentially contemplate if they did not potentially phrase the independent claims too restrictively so that they lost scope by not first testing a more general claim version.}. Previous works named this task ``patent acceptance prediction'' \citep{suzgun2022harvard}, but we correct it to ``granting prediction'' in this paper because patents are granted at best but not accepted. Such an automated method could support patent examiners, patent applicants, and external investors. Due to the complexity of patent examination, this process typically requires a long time. Some cases have still not achieved the final decision after two decades, when they typically expire at last. Thus, the automation of this process can help patent offices manage their workload more efficiently. If automated examination procedures achieve to be bias-free, artificial intelligence could reduce the existing examiner dependence and also likely speed up the procedure. Furthermore, such automated examination promises to increase the quality of examination, particularly its existing bias and variability. The quality of examination has been and is such a severe issue that it, for instance, led to the America Invents Act of 2011. In addition, patent applicants can gain valuable insight, be less exposed to bias and individuality of patent examiners, and improve the invention based on the prediction of the outcome. NLP may reduce the excessive cost, particularly for small companies, to participate in the intellectual property system. Currently, drafting applications and sending them for examination can generate overwhelming costs of thousands of dollars every round for small businesses without internal resources. Such bills further increase if an examiner shows little support and a hearing needs to be arranged in the process. Thus, automation of the intellectual property process has the chance to stimulate technology development and increase the overall technological competitiveness of a society. Similarly, investors can make better-informed decisions and create strategic plans according to the predicted outcome of patent applications. 

\subsubsection{Methodologies for Granting Prediction}
Only few studies have investigated grating prediction, probably because of the difficulty of the task. \citet{suzgun2022harvard} provided baselines for patent granting prediction by training various models on patent abstracts and claims, such as convolutional neural networks (CNNs) and BERT. None of the accuracy was higher than 64\%, indicating a clear need for improvement. In addition, \citet{jiang2023deep} proposed the PARCEL framework, which combines text data and metadata for patent granting prediction. PARCEL learns text representations from patent abstracts and claims by using the contextual bidirectional long short-term memory (LSTM) method and captures context information based on active heterogeneous network embedding (ActiveHNE) \citep{chen2019activehne}. The authors incorporated attention mechanisms when combining text and network attributes for final prediction. The results showed that this method could achieve approximately 75\% accuracy, which exceeded many previous models, such as PatentBERT \citep{lee2020patent}. It is important to note that the datasets used in the above two studies are different, so the results are not comparable directly. Nonetheless, the studies published their datasets for future research on related topics. In summary, the prediction performance in any of the reported settings is far from satisfactory. Due to the high formality, intensive future research on this complex topic promises improvement. 

\begin{figure}[!htbp]
    \centering
    \includegraphics[width=.49\textwidth]{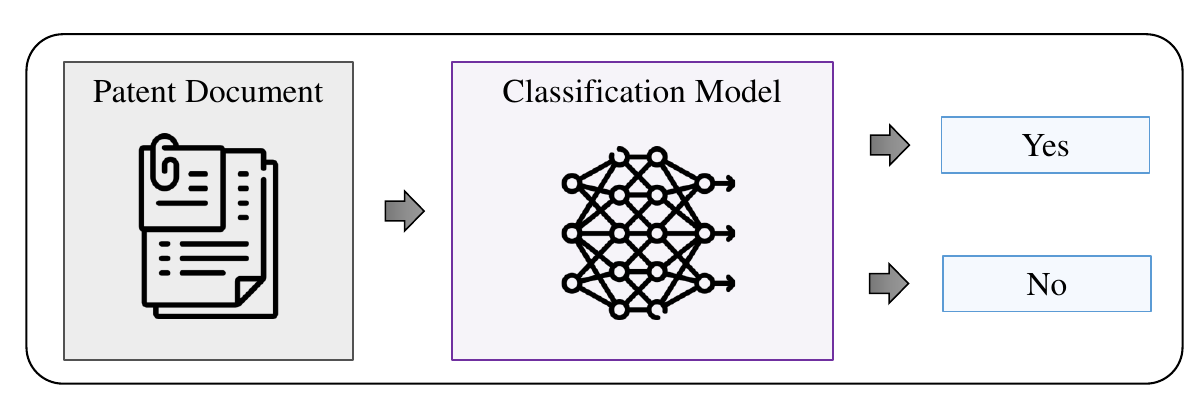}
    \caption[]{Pipeline of patent granting prediction and litigation prediction }
    \label{fig:accept}
\end{figure}

\subsection{Litigation Prediction}
\label{litigationprediction}
\subsubsection{Task Definition of Litigation Prediction}
Litigation prediction aims to predict whether a patent will cause litigation, typically between two companies. Patent litigation involves various types. For example, one of the most common scenarios in patent litigation is infringement, where patent holders claim that their patent rights are infringed. Additionally, third parties can question the patent validity, such as pointing out prior art that can prove the patent's claims are not novel and/or inventive.\footnote{Strategically, most defendants immediately trigger cross action going after the validity of the respective rights when they receive an infringement indictment when the case is not rock solid.} An early identification of whether a patent might become the subject of litigation can help companies manage the quality of their patent portfolio \citep{wu2023multi}. 

\subsubsection{Methodologies for Litigation Prediction}
Manual litigation prediction is an expensive and time-consuming task faced by companies, which is based on intensive reading and experience, and typically does not exceed mere speculation. Hence, researchers developed different methods to automate the process and reduce costs. \citet{campbell2016predicting} adopted multiple pieces of patent information to predict litigation with machine learning models. The authors combined metadata (e.g., number of claims, patent type), claim texts, and graph features that summarized patent citations to inform their predictions. The results demonstrated that the metadata has the highest impact on litigation prediction. Apart from using patent contents, such as metadata and texts, \citet{liu2018patent} incorporated lawsuit records of companies to collaboratively predict litigation between companies. The authors trained deep learning models that integrated patent content feature vectors and tensors for lawsuit records. This method brought more than 10\% gain in precision compared to using text inputs only. Moreover, \citet{wu2023multi} improved the numerical representations of patents and companies to capture the complex relations among plaintiffs, defendants, and patents more effectively. The authors demonstrated that this method could increase the performance by up to 10\% compared to the previous approach and show robustness under various data-sparse situations.

\subsection{Patent Valuation}
\label{valuation}
\subsubsection{Task Definition of Patent Valuation}
The value of a single patent, patent applications, or an entire portfolio is highly important for young companies, where investment relies. However, the value of patents is not obvious and can reflect expectations as most immaterial assets do. Firstly, the single patents of a portfolio can have a lower value than the portfolio. Thus, patents owned by a third party may influence the value of a patent, e.g., if one depends on the other or if they solve the same problem with alternative solutions, i.e., feature sets.  Furthermore, the value may depend on the function the specific patent(s) are supposed to fulfill and therefore also be different depending on the (potential) owner.

For blocking competitors or enforcing licensing deals, a single patent of a portfolio can be highly valuable if all the features of an independent claim are unavoidable. A single patent can thus turn into a road block for others. If, however, a patent application for the same purpose is significantly tattered during the examination that the independent claim requires substantial avoidable features, the patent is granted but without much value.

However, for the freedom-to-operate, patents should not contain dependencies, which means that active patents owned by others would need to be used to exploit the invention. Thus, in this case, the patent has only value in combination with rights to previous patents. Furthermore, the portfolio should ideally not leave too many gaps where others could establish themselves. In all those cases, however, the scope of the patent, i.e., how wide of a range the independent claims in the granted state after examination can cover, is an essential factor determining the value, although not the only one.

Early identification of high-value patents can help various industries and stakeholders establish optimal strategies. This task can be either a regression or classification problem. The regression task aims to predict the actual number of the value. For example, \citet{lin2018patent}  used the number of citations as a measure of patent quality, because previous works have indicated that the number of citations is closely related to the patent value \citep{albert1991direct, harhoff1999citation}. However, the globally existing correlation between citations and patent value appears to include large variability, so it may not be the best individual predictor. 
More importantly, the citations typically trickle in after years, often when an assessment of the value of a patent is no longer needed. Other simpler and more accurate indicators may be already available, such as assigned revenue share in actual products, licensing fees, or sale of the patents.

In the context of binary classification, the purpose is to predict whether a patent is of high quality \citep{hu2023evaluation}. Additionally, researchers classify the value of patents into different levels. For example, \citet{zhang2024research} graded the quality of patents into four classes based on expert consultation and questionnaire surveys. 

\begin{figure}[!htbp]
    \centering
    \includegraphics[width=.49\textwidth]{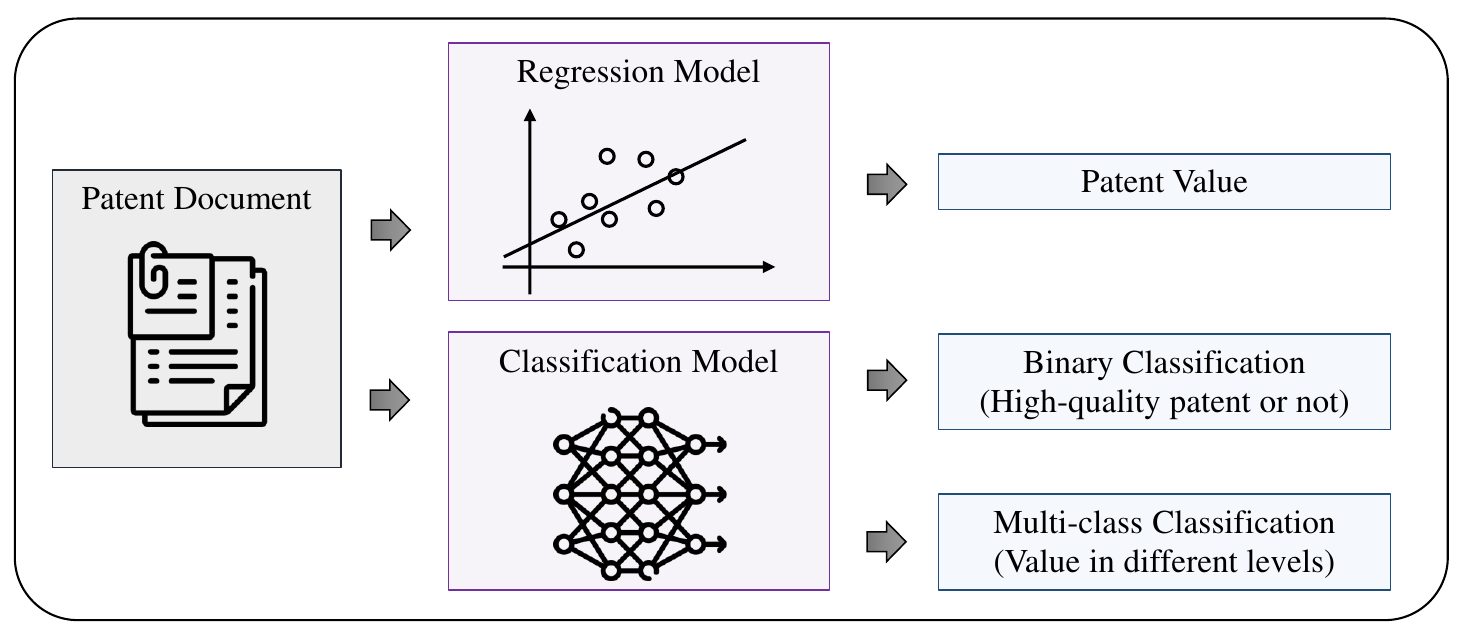}
    \caption[]{Specifics of patent valuation tasks }
    \label{fig:valuation}
\end{figure}

\subsubsection{Methodologies for Patent Valuation}
Previous studies usually adopted pre-defined indicators for patent valuation. \citet{hu2023evaluation} designed four dimensions to evaluate patent quality, including legal value, technological value, competitiveness value, and scientific value. The authors extracted and transformed the indicators into numerical values for training machine learning models. Additionally, \citet{du2021personalized} incorporated patent inventors’ reputations as a new indicator and proved the effectiveness for patent quality evaluation. 

Text data are also important for patent valuation. Researchers applied NLP to extract semantic features of patent texts for quality assessment, which contained detailed contextual information \citep{lin2018patent, chung2020early}. \citet{lin2018patent} used attention-based CNNs to extract semantic representation from patent texts, and adopted GNNs to represent patent citation networks and attributes. These feature vectors were concatenated to predict the quality of new patents. This method outperformed baseline approaches, such as using CNNs alone. \citet{chung2020early}, on the other hand, used both CNNs and bidirectional LSTM for patent valuation classification based on abstracts and claims. The authors transformed texts into word embeddings as inputs and classified patents into three quality levels for prediction. The model achieved over 75\% precision and recall in identifying high-value semiconductor patents. 

Moreover, \citet{liu2023multi} proposed a multi-task learning method that jointly trained classification models for the identification of high-value patents and standard-essential patents, i.e., patents that are required to implement a specific industry standard. These two tasks were related, which made the collaborative training effective. The results showed that the average performance improvement was around 1--2\% compared to single-task learning.

\subsection{Technology Forecasting}
\label{technologyforecast}
Technology forecasting refers to predicting future technological developments based on existing patent data. Many businesses, researchers, and policymakers consider it important to understand the direction of technological innovation, design strategies, and adjust their investment. 
As technology forecasting is a broad concept, we summarize and categorize existing studies into three sub-types, including emerging technology forecasting, technology life cycle prediction, and patent application trend prediction. 

\begin{figure*}[!htbp]
    \centering
    \includegraphics[width=.8\textwidth]{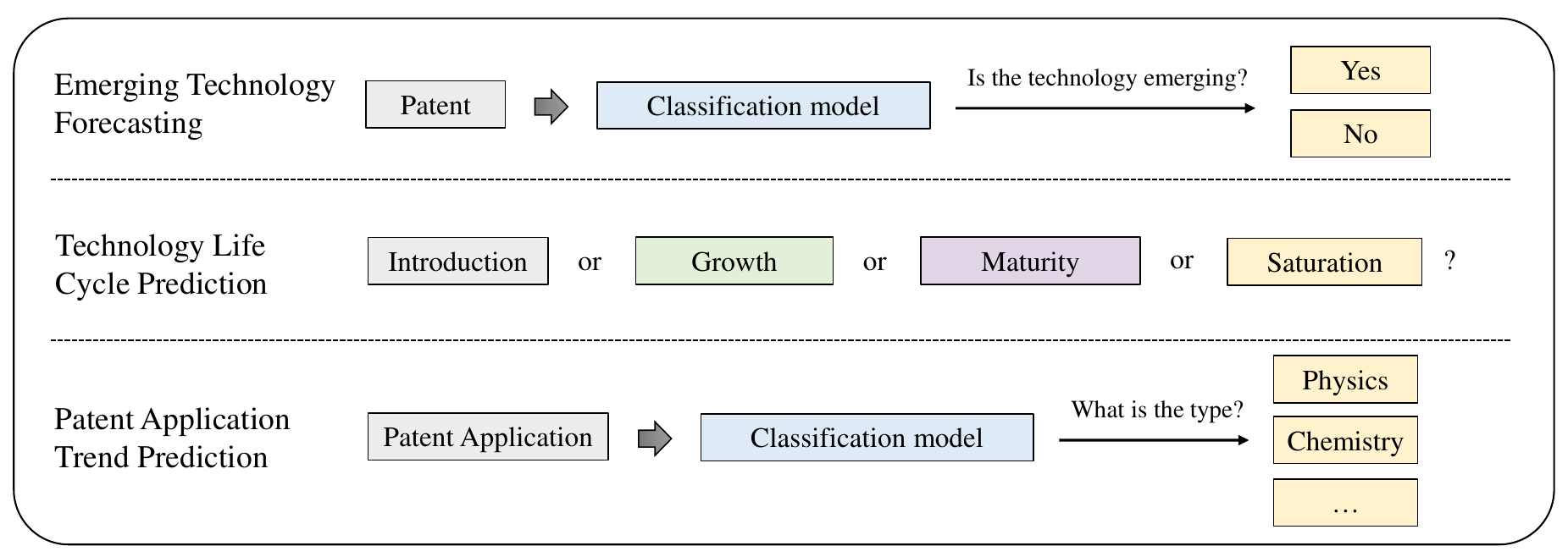}
    \caption[]{Specifics of technology forecasting tasks }
    \label{fig:forecast}
\end{figure*}

\subsubsection{Emerging Technology Forecasting}
Emerging technology forecasting is a binary classification task, which predicts whether a patent contains emerging technologies. This task may share similarities with some forms of patent valuation because high-value patents are more likely to become emerging.  Due to the similarity with valuation, previous work identified emerging technologies by assessing the quality of patents \citep{lee2018early}. In addition, studies developed more elaborate machine-learning models for this task. \citet{kyebambe2017forecasting} labeled a cluster of patents as emerging technology if it demonstrated evidence of possessing features that could introduce a new technological class shortly. The authors extracted patent information, such as patent class and number of citations, and trained various models (e.g., support vector machine, naive Bayes) that achieved approximately 70\% precision. 

As machine learning methods usually require large datasets for training, \citet{zhou2020forecasting} aimed to evade this issue with a generative adversarial network (GAN). The authors labeled emerging technologies based on Gartner’s emerging technology hype cycles (GETHC), which aims to describe a specific stage of development of an emerging technology. They selected and extracted some patent attributes, such as the number of backward citations, to train machine learning models for prediction. Importantly, they incorporated a GAN to generate synthetic samples in the form of feature vectors for training. The synthetic data finally improved the prediction precision to 77\% on the real test set.

While the above studies investigated emerging technologies from a technological perspective, researchers also evaluated social impacts based on related website articles \citep{zhou2021deep}. The authors measure social impact through the frequency of patent-related keywords found in website articles, where ``website articles'' encompass all articles on emerging technology-related websites, and ``patent keywords'' are non-virtual words present in patent titles. They collected a multi-source dataset, including 129,694 patents and 35,940 website articles. They selected 11 patent indicators as inputs and classified social impacts into three levels as labels. Subsequently, they trained a deep learning model that achieved 74\% accuracy.

\subsubsection{Technology Life Cycle Prediction}
The aim of technology life cycle prediction is to estimate the stages of a particular technology from its inception to the eventual decline. In established classifications, the typical stages of a technology life cycle include introduction, growth, maturity, and saturation \citep{mansouri2023determining}. The technology life cycle and the related technology adoption (as well as the hype cycle) are important aspects of innovation and product management. 

The first stage involves the conception and initial development of the technology. If the technology proves viable, it will be adopted and developed rapidly, with an increase in market acceptance. As the technology becomes widely adopted, it enters a maturity stage. Finally, in the saturation stage, the previous technology will be replaced by a new one soon. Understanding and predicting the life cycle of a technology is a key part of strategic planning and market analysis, especially in fast-evolving fields like artificial intelligence.
The phase of a technology affects the types of problems for a manufacturer (e.g., maturity, field experience, not established design rules as well as design optima, and reliability in early stages vs.\ mere cost focus later on), the type of customers (smaller customer segment of high-tech enthusiasts and soon early adopters vs.\ the later mass market), and also the company strategy (low volume, high-profit high-tech companies vs.\ mass-market).

Previous studies used a hidden Markov model (HMM) for technology life cycle prediction \citep{lee2016stochastic, mansouri2023determining}. Researchers extracted several indicators from patents, such as patent class and citations, and investigated the state transferring probabilities. They trained the hidden Markov model to predict the probability of each phase in the future. However, using intrinsic indicators from patents as the sole predictor of technology life cycles may not be sufficient. The adoption and success of patents are heavily influenced by social needs and user acceptance. In addition, other extrinsic factors, such as market trends, competitive landscape, and strategic decisions, can influence the pace of technology adoption and development. Therefore, a multi-dimensional approach that combines intrinsic patent data with extrinsic information will likely provide a more accurate and holistic prediction of technology life cycles.

\begin{figure*}[!htbp]
    \centering
    \includegraphics[width=\textwidth]{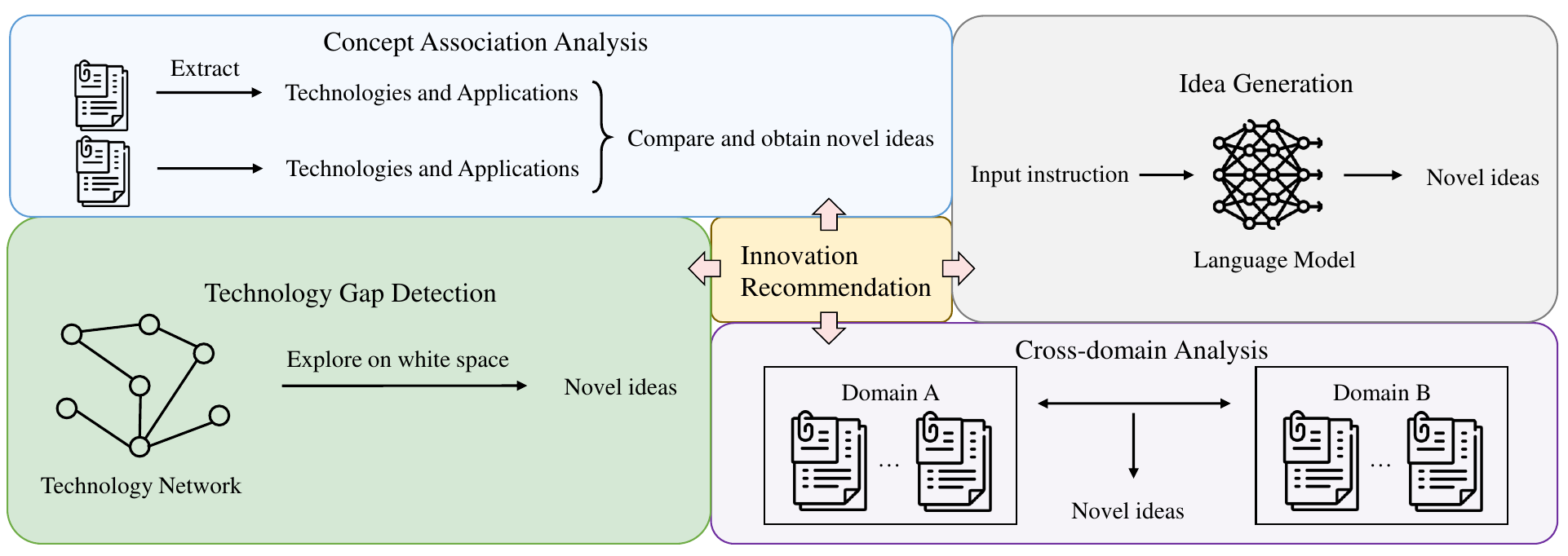}
    \caption[]{Methods of innovation recommendation task}
    \label{fig:innovation}
\end{figure*}

\subsubsection{Patent Application Trend Prediction}
Patent application trend prediction aims to predict patent classification codes for which a company will apply in the future, given a sequence of patents that were previously applied by the same company. Precise forecasts of trends in patent applications can enable businesses to devise effective development strategies and identify potential partners or competitors in advance \citep{zou2023event}. Researchers implemented and tested dynamic graph representation learning methods for this task under various experimental conditions to demonstrate the effectiveness \citep{zou2023event}. Although the authors significantly improved the performance compared to machine learning baselines, the recall was hardly higher than 20\% on overall experiments, suggesting substantial room for improvement. Company-dependent patent application prediction is a newly proposed task and only few papers have investigated solutions.

\subsection{Innovation Recommendation}
\label{innovationreccomendation}
\subsubsection{Task Definition of Innovation Recommendation}

Innovation recommendation refers to the process of suggesting new ideas or methods in a technology context. The automation of this task can enhance competitiveness, efficiency, or user experience. Innovation recommendations should help companies to identify novel research and development opportunities.\footnote{
Strictly speaking, the use of AI to generate inventions may raise legal issues, notably around inventorship and ownership. The use of AI for writing a patent could do so if the AI starts to add information or hallucinate so that it could be considered inventive, not considering copyright and authorship rights.   Traditionally, patents are granted to human inventors, i.e., natural persons. 
The US Patent and Trademark Office, at present, does not consider AI-generated inventions as patentable, which limits how technology gap detection and idea generation can be used to create patentable inventions. Practically, natural co-inventors might suppress the co-inventorship of an AI until technical means are available to detect AI contributions. No precedence seems available if that would collide the inventors' oath filed with the application or would set an inventor into bad faith.
These issues highlight the need for evolving legal standards to accommodate the rapid advancements of AI technology in the patent system.}

\begin{figure*}[!htbp]
    \centering
    \includegraphics[width=\textwidth]{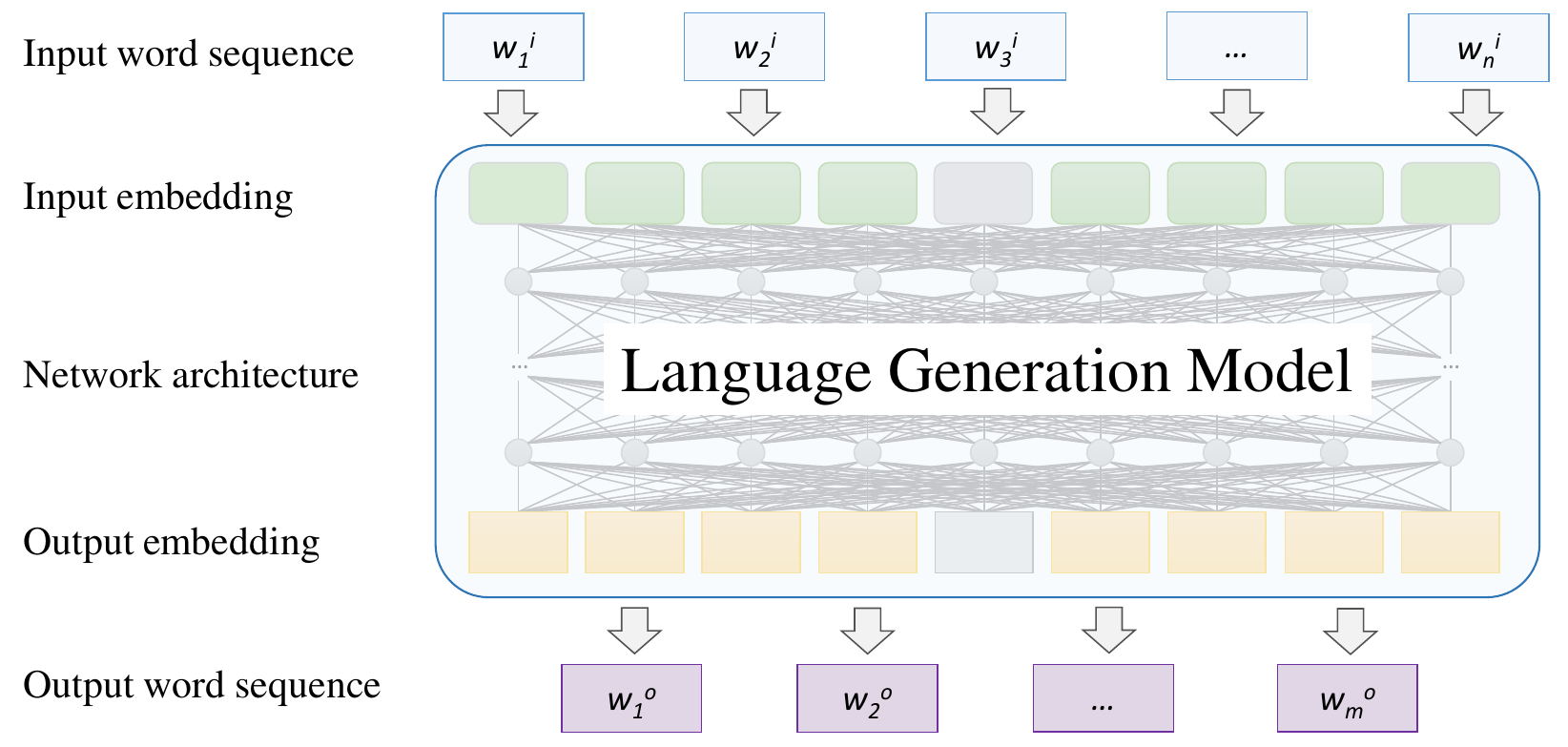}
    \caption[]{Illustration of text generation tasks }
    \label{fig:generation}
\end{figure*}

\subsubsection{Methodologies for Innovation Recommendation} 

\textbf{Concept Association Analysis. } Researchers used NLP techniques to identify key concepts in patent documents and analyze their associations to discover novel combinations of concepts and suggest potential directions for innovation. For example, \citet{song2017discovering} identified technologies with technical attributes similar to a target technology. Thus, the authors could obtain novel technology ideas by applying these similar but distinct technical attributes to the target technology. From a patent-law perspective, it would be naturally interesting whether and how such technology ideas were synthesized out of existing patents. 
Potentially, the inclusion of intentional variability through \textit{higher temperature} may lead to a more creative tool.

\textbf{Technology Gap Detection. } Technology gap detection algorithmically searches for areas within a technology field that are underdeveloped or unexplored. Such technology gaps can indicate new opportunities for research and innovation. TechNet is a large-scale network consisting of technical terms retrieved from patent texts and associated according to pair-wise semantic distances~\citep{sarica2020technet}.  \citet{sarica2021idea} focused on white space surrounding a focal design domain and adopted semantic distance to guide the inference of new technical concepts in TechNet. However, the strong dependency of TechNet on the terminology in patents may need further research because quite a number of patents form their own world of terminology and technical terms.

\textbf{Cross-Domain Analysis. } Cross-domain analysis links patents across different technological fields to find proper cross-domain techniques. Such cross-domain innovations can sometimes lead to breakthrough developments. \citet{wang2023discovering} applied causal extraction with a BERT-derived model to patents from different domains to output cause--effect matches to represent technology--application relationships that can undergo similarity comparison afterwards. The results are connections between patents from different fields that appear to have similarities in causality, presumably technology--application links, and may be good candidates to stimulate each other's field. It supports cross-domain comparisons of technologies and applications to identify prospects for multiple applications of a specific technology. However, this approach may depend strongly on the terminology used in both patent texts.

\textbf{Idea Generation. } Researchers developed generative language models to output innovative ideas automatically. A previous study fine-tuned the GPT-2 model to generate idea titles based on input keywords in a specific domain \citep{zhu2022generative}. The authors collected patent titles and extracted keywords from them. They used keywords as inputs and fine-tuned GPT-2 to generate titles. However, patent titles are often generic for legal reasons. For example, ``charging station and methods to control it'' covers almost any conceivable detail of a car charger.

\section{Patent Generation Tasks}
\label{generationtask}

Patent text generation tasks aim to automatically create coherent and contextually relevant texts based on input prompts or requirements. As shown in Fig.~\ref{fig:generation}, language generation is a sequence-to-sequence mapping of texts. The input and output are both word sequences, $x=[w_1^i, w_2^i, \ldots, w_n^i]$ and $y=[w_1^o, w_2^o, \ldots, w_m^o]$. There are four types of patent text generation tasks according to different objectives, including summarization (Section~\ref{summarization}), translation (Section~\ref{translation}), simplification (Section~\ref{simplification}), and patent writing (Section~\ref{patentwriting}).

\begin{table*}[t!]
\centering
\footnotesize

\begin{tabular}{lcccc}
\toprule
Model & Architecture & R-1 & R-2 & R-L \\
\midrule
Seq2Seq \citep{sharma2019bigpatent} & LSTM & 28.74 & 7.87 & 24.66 \\
SentRewriting \citep{sharma2019bigpatent} & LSTM & 37.12 & 11.87 & 32.45 \\
PEGASUS \citep{zhang2020pegasus} & Transformer & 43.55 & 20.43 & 31.80 \\
BigBird-PEGASUS \citep{zaheer2020big} & Transformer & 60.64 & 42.46 & 50.01 \\
LongT5 \citep{guo2022longt5} & Transformer & 76.87 & 66.06 & 70.76 \\
\bottomrule
\end{tabular}
\caption{Comparison of representative models on the BigPatent benchmark \citep{sharma2019bigpatent} for patent summarization as measured by ROUGE score \citep{lin2004rouge}}
\label{tab:bigpatent_results}
\end{table*}

\subsection{Summarization}
\label{summarization}
\subsubsection{Task Definition of Summarization}

Patent summarization aims to create concise and informative summaries of patent documents. Given the complex and technical nature of patents, summarization helps in extracting the most important information and paraphrasing in more accessible language depending on the audience. The target audience may be as diverse as patent examiners, researchers, development engineers, and legal professionals. Additionally, summarization can extract the gist of patent subgroups within classification systems, which is essential for organizing and managing large patent datasets \citep{souza2021comparative}. The summarization process involves understanding and condensing the key aspects of a patent, such as claims, background, and detailed descriptions of the invention.

\subsubsection{Methodologies for Summarization}

Extractive and abstractive summarization are two primary approaches for text summarization.

\noindent\textbf{Extractive Summarization. } Extractive methods refer to selecting and extracting key phrases or sentences directly from the text of the patent. The goal is to retain the most significant and representative parts of the original document without altering the text. For instance, \cite{souza2019using} used different approaches, such as term frequency-inverse document frequency (TF-IDF) and latent semantic analysis (LSA) \citep{deerwester1990indexing}, to choose the most representative sentence for each classification subgroup\footnote{Patent sub-groups are the most specific level in the patent classification hierarchy, see Section~\ref{class_scheme}}. 

Compared to abstractive summarization, extractive methods are simpler to implement, as they do not require complex language generation capabilities. In addition, since the extracted sentences are taken verbatim, the original context, meaning, and particularly the wording of texts are well-preserved.
However, this method may lack coherence, because extractive summaries can sometimes be disjointed. Moreover, it is hard to capture the essence of the text if the key information is not explicitly stated. In that sense, the summary can have the same limitations as the claims, which only contain the essence of the invention (in features) but are often poorly intelligible without the description as their dictionary.

\noindent\textbf{Abstractive Summarization. } Abstractive summarization involves rephrasing original documents into shorter texts that retain essential information, a process demanding a deeper understanding of the text and the ability to generate coherent summaries. Researchers commonly use sequence-to-sequence models for this task, where the models receive original texts as inputs and produce summarized texts as outputs. 

\cite{trappey2020intelligent} trained a summarization model based on long short-term memory (LSTM), i.e., a sequential deep learning network, with attention mechanisms to extract essential knowledge from patent documents. Although promising, the performance was surpassed by the recently proposed transformer-based models. Concretely, fine-tuning transformer-based language models for summarization has shown remarkable performance in both general and patent domains \citep{zhang2020pegasus, zaheer2020big}. Whereas the focus of most research has been on single patent summarization, \cite{kim2022multi} introduced a method for multi-document summarization. The authors showed that their models enabled high-quality information summary for a large number of patent documents, which could be used to facilitate the activities of researchers and decision-makers. 

Abstractive summarization can produce more concise, fluent, and cohesive summaries, by condensing and synthesizing information from multiple parts of the source text. Nonetheless, there is a higher risk of distorting facts, confusing meanings of terms that may deviate from everyday language on which the embedding of language models was trained, or misrepresenting the original text because of the involvement of paraphrasing and rewording. Additionally, abstractive summarization can be resource-intensive in terms of computational power and training data. 

Table~\ref{tab:bigpatent_results} compares representative models on the BigPatent benchmark \citep{sharma2019bigpatent} for patent summarization. Transformer-based models outperform previous LSTM-based methods. In addition, models with long input lengths, such as LongT5 \citep{guo2022longt5}, significantly increase the performance, because patent descriptions usually contain more than 10,000 tokens \citep{suzgun2022harvard}.

LLMs have achieved satisfactory performance on most summarization tasks in the general domain, even surpassing the benchmark of reference summaries \citep{pu2023summarization}. The state-of-the-art method for text summarization is to use LLMs for abstractive summarization \citep{rao2024single}. However, how these LLMs, such as GPT-4, perform on patent summarization is still worth investigating. The technical content and legal information in patents may pose some challenges to LLMs compared to normal summarization.

\subsection{Translation}
\label{translation}
\subsubsection{Task Definition of Translation}

Patent translation refers to the process of converting patent texts from one language to another, which is crucial in the global landscape of intellectual property. Translation can make patents accessible to individuals, companies, and researchers across different linguistic regions, fostering cross-border innovation and cooperation. In addition, translation can ensure that patents comply with the legal requirements of different countries and obtain protection under the law. 

Automated translation is maybe one of the earliest language-processing applications in the intellectual property domain and has become a highly established tool in the field already, which is used by patent offices routinely. Patent offices can leverage the large body of meticulously translated documents in many languages for model training. 
For instance, the European Patent Office (EPO) described their approach to identify parallel patent documents and sentences to generate training and evaluation datasets for patent translation \citep{wirth2023building}. The EPO offers machine translation on its free database interface, Espacenet. The engine in the back, called Patent Translate, was initially developed in 2012 in collaboration with Google. It initially started with the most frequent six languages and enhanced to now understand all of the office's languages in addition to many others, such as Chinese, Japanese, and Korean. 
At a similar time, the World International Patent Office (WIPO) introduced a machine translation tool, WIPO Translate, integrated into their search engine Patentscope. It was reported to outperform standard machine translators for everyday texts, such as Microsoft Translate \citep{pouliquen2015full}. Standard translators tend to vary terms and use synonyms in translations, which are acceptable or even desired in everyday language. 
However, the substitution is not acceptable and can render a patent worthless when it involves features or feature-relevant language. Therefore, in designing generative models, the tendency to introduce variations, often referred to as the temperature in LLMs, must be carefully managed.

\subsubsection{Methodologies for Translation}

Machine translation has been widely studied and gained significant progress in the general domain, such as statistical methods and neural machine translation \citep{wang2022progress}. However, these models may not achieve strong performance in patent translation, because patents contain highly specialized technical jargon and precise terminology, challenging accurate translation. Therefore, researchers focused on the adaptation of general machine translation models to the patent domain. 

\noindent\textbf{Patent Translation Datasets. } A feasible approach is to train the translation model on the above-mentioned multi-language patent data. For example, \cite{heafield2022europat} proposed the EuroPat corpus, which includes patent-specific parallel data for official European languages: English, German, Spanish, French, Croatian, Norwegian, and Polish. In addition, the European Patent Office described their approach to identify parallel patent documents and sentences to generate training and evaluation datasets for patent translation \citep{wirth2023building}. 

\noindent\textbf{Fixing Specific Errors. } Since accurate language is crucial in patent texts while language models are even designed to vary based on distance metrics of word meanings, for instance, research has studied pre- and post-processing methods for fixing common problems. As an example for terminology errors, \cite{ying2021errors} identified eight types of terminology errors in patent translation from English into Chinese and suggested solving these errors by pre-editing the source texts. In addition, \cite{larroyed2023redefining} compared ChatGPT \citep{ray2023chatgpt} and the Patent Translate system of the European Patent Office on patent translation task. In their analysis, Patent Translate could exploit its training on patent corpora and specialization in patent translation to outperform ChatGPT in accuracy. ChatGPT, however, performed better on language structure and overall textual coherence. In linguistics, these two qualities might refer to paradigmatic versus syntagmatic properties. Such results suggest that a combination of both advantages may open new opportunities. One part of the solution may be on the model side, where transformers or a combination of recurrent neural networks (RNNs) and transformers may improve structural, i.e., syntagmatic shortcomings. Further deriving a domain-specific model through training and/or fine-tuning to patent corpora promises to enhance both syntagmatic and paradigmatic performance through the incorporation of more patent-typical language structures as well as sharper control over terminology-related issues. 

While translation systems, such as Patent Translate, offer valuable tools for understanding patent documents in various languages, it is important to recognize that machine translations may not always achieve the precision required for legal or formal purposes. Therefore, for critical applications such as patent filings or legal proceedings, consultation with professional human translators is advisable to ensure the highest level of accuracy. From the above studies, we can conclude that training specific LLMs on extensive patent translation datasets is highly promising to further improve translation accuracy and language precision.

\subsection{Simplification}
\label{simplification}
\subsubsection{Task Definition of Simplification}

Patent simplification refers to the process of translating complex and technical patent documents into more straightforward readable language. The difficulty of patent texts is often two-fold: First, the content is supposed to be at the forefront of science and technology and/or refer to particular details that require an exceptionally good understanding of the context. Second, the language that describes the invention focuses on precision, accuracy, and sometimes intentional flexibility to avoid unnecessary reduction of the scope instead of readability. This precision typically leads to high repetitiveness in both terminology and structure of sentences, paragraphs, and sections. Especially, sentences are often overburdened with specifications for precision (typically relative or adverbial clauses) or examples and alternatives for a wide enough scope.

Different from patent summarization, all information should ideally be retained in simplified texts in the task of patent simplification. The aim is to improve readability and make patent texts more accessible to a wider audience, such as technical experts without an interest in legal language. Simplification can involve multiple aspects, such as summarizing key concepts, removing jargon, and rephrasing technical terms in layman's terms.

\subsubsection{Methodologies for Simplification}

\begin{figure*}[!htbp]
    \centering
    \includegraphics[width=\textwidth]{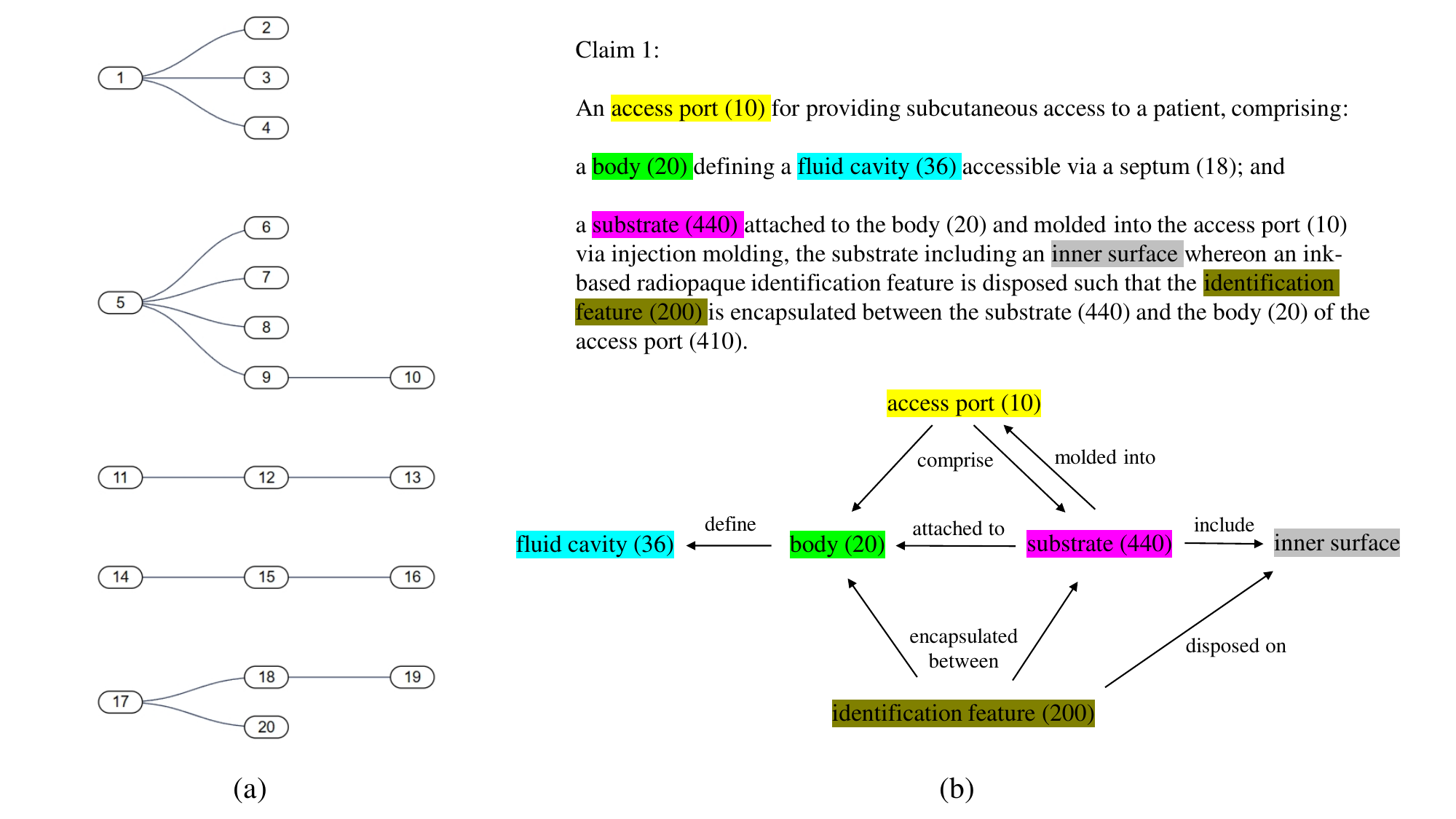}
    \caption[]{(a) The claims tree of EP~2\,{}346\,{}553 (specifically version A1 of it, i.e., the not granted (therefore A and not B, C or similar) first (accordingly counter 1) filing of the applicant), which can be generated from the text based on the logical relationship between the 20 claims. (b) Claim~1 of patent EP~2\,{}346\,{}553 B1. Each claim can further be split into features, e.g., all elements that a device collectively has to contain to be considered an embodiment of the invention. }
    \label{fig:claimtree}
\end{figure*}

\textbf{Formatting. }Original documents undergo formatting to enhance accessibility without modifying the texts. Patent offices, for instance, often structure claims in their interfaces. On the macro level, the dependency on claims can be automatically derived from the claims (e.g., \textit{Invention of one of the previous claims, wherein \ldots}) to form a tree structure of claims. An example claims tree is shown in Fig.~\ref{fig:claimtree}. Within a claim, the individual features can be separated so that a graphical representation can indicate which features get added by each claim. The independent claims may further have a two-part structure (e.g., European Patent Regulations Rule 43, PCT Regulations Rule 6.3, US 37 CFR 1.75(e)) where the first part (preamble) lists the features of the prior art and the second part (typically introduced with an adverb as \textit{whereby}, a participle as \textit{comprising} or \textit{characterized by}, or a relative pronoun) refines it with further features ({characterizing portion}).

While the European Patent Office uses explicit references in the text to other claims to generate claim trees, such as \textit{according to claim~1}, \cite{andersson2013domain} developed a dependency claim graph to exploit implicit references by detecting discourse references. The authors defined rules to extract and use linguistic information to build the tree, including part-of-speech, phrase boundaries, and discourse theory. The claim tree simplified the connection and relation between complex claims to improve patent accessibility. In addition, \cite{sheremetyeva2014automatic} designed two levels of visualization of patent documents. The macro-level simplification involved the development of claim trees based on pre-defined rules. Meanwhile, micro-level simplification used both rule-based and statistical techniques that rely on linguistic knowledge. The authors highlighted nominal terms with the reference in patent description, aiming to improve readability by providing terminology at a glance. They also segmented complex claims into simple sentences to increase accessibility. This hierarchy visualization system effectively increases the overall productivity in processing patent documents.

To simplify patent claim texts, \cite{ferraro2014segmentation} proposed some rules to segment the original claims into three components, consisting of the preamble, transitional phrase, and body. Although the clearer presentation of patent claims improved the readability and accessibility, the original complicated texts were not modified. Thus, it could still be challenging to read and understand the technical documents. 

\noindent\textbf{Paraphrasing. } Paraphrasing refers to rewriting the original complex sentences into simpler versions with the text meaning unchanged. \cite{kang2018text} proposed a simplification system based on the analysis of patent syntactic and lexical patterns. This simplification system could detect complex sentences and simplify those sentences through splitting, dropping, and modifying. Paraphrasing can break down complex concepts into simpler language to be accessible to a broader audience. Nonetheless, paraphrasing may lose subtle nuances and specific terminologies, which are crucial in some contexts. 

Additionally, some studies focused on sequence-to-sequence models for text simplification in the general domain. For example, \cite{martin2020controllable} proposed to use explicit control tokens for attributes, such as length, lexical complexity, and syntactic complexity, to tailor simplifications for the needs of different audiences.  However, these models were rarely used for patent simplification because of the lack of in-domain datasets. \cite{casola2023creating} aimed to break the boundary by proposing data generation methods for patent simplification. The authors used a Pegasus-based---a transformer-based model for abstractive summarization \citep{zhang2020pegasus}---paraphraser trained on general-domain datasets in a zero-shot fashion to obtain simplification candidates from complex patent sentences. Subsequently, the authors discussed filtering to select appropriate candidate patent documents for simplification only and obtained the first large-scale dataset for patent sentence simplification. 

In contrast to well-established tasks of patent summarization and translation, patent simplification is less investigated and worth more attention. Particularly, established benchmarks to compare different models and approaches are important to accelerate the development of this topic. LLMs appear an ideal tool for this task, despite the challenges of processing complex technical documents with legal requirements.

\begin{table*}[!t]
\centering
\footnotesize
\begin{tabular}{p{.33\textwidth}p{.25\textwidth}p{.3\textwidth}}
\toprule
Task   & \multicolumn{1}{c}{Method}    & \multicolumn{1}{c}{Data} \\ \midrule
Claim generation \citep{lee2020patent}   & Fine-tunning GPT-2   & 555,890 patent claims from granted U.S. utility patents in 2013 (Claims are the first and independent claims) \\
Title, abstract, and claim generation \citep{lee2020controlling}  & Fine-tuning transformer-based models  & 404,975,822 utility patent from 1976 to 2017-08 in Google Patents Public Datasets \\

Description-based claim generation \citep{jiang2024can} & Fine-tuning Llama-3-8B, GPT-4 (zero-shot) & 9,500 pairs of patent descriptions and claims \\

Claim generation, text infilling  \citep{christofidellis2022pgt} & Fine-tuning GPT-2 with Multi-task Learning & 11,600,000 patents published between 1998 and 2020  \\

Claim revision \citep{jiang2024patent} & Fine-tuning Llama-3.1-8B, GPT-4 (zero-shot) & 22.6K pairs of application-level and granted versions of patent claims \\

Specification generation \citep{wang2024patentformer} & Fine-tuning GPT-J and T5 & 13,725 patents from USPTO in CPC code of G06N \\
Paper to patent generation \citep{knappich2024pap2pat} & Fine-tuning Llama-3-8B & 1,800 pairs of academic papers and corresponding patent documents (outlines) \\ 

\bottomrule
\end{tabular}
\caption{Previous studies for patent writing task. }
\label{text_generation}
\end{table*}

\subsection{Patent Writing}
\label{patentwriting}
\subsubsection{Task Definition of Patent Writing}
Patent writing refers to the automated creation of patent texts, including various sections such as abstracts, claims, and descriptions. The primary goal of patent text generation is to assist inventors, patent agents, and attorneys in drafting patent applications more efficiently and effectively. Text generation is a difficult task, especially in the patent domain due to the linguistic, technical, and legal complexities. These facts pose significant challenges for language models to handle complex legal jargon and terminology while ensuring accuracy and adherence to technical norms.

\subsubsection{Methodologies for Patent Writing}
A few studies have investigated patent writing for different sections based on fine-tuning LLMs. \citet{lee2020patentgenerate} provided an initial proof-of-concept study for patent claim generation. The authors fine-tuned the GPT-2 model to generate patent claims, but the quality of generated claims was not measured, which reduced the actual value of the work. In further research, \citet{lee2020controlling} trained the GPT-2 model to map one section to another, such as generating abstracts based on title and generating claims from abstracts. 
Nonetheless, patent professionals may raise concerns about the validity of the tasks. Patent titles only include a few words and lack specificity, which makes title-based generation unfounded. In contrast, abstract-based claim generation sometimes makes sense because the abstract in some patents is just a paraphrased version of the first independent claim (without the legal phraseology). Hence, this task is straightforward by extracting and revising the abstract. However, abstracts are typically crafted to be general and reveal minimal details of the invention, adhering to requirements of patent offices \citep{epo2020, wipo2020pct}. Consequently, it is almost impossible to generate detailed invention features required by claims based solely on general and vain abstracts. Therefore, both title-based and abstract-based generation tasks are not well-conditioned. 
Alternatively, patent descriptions include all details and specific embodiments of the invention, which makes description-based patent generation notably promising for future research.
Recent research constructed a dataset with 9,500 examples and evaluated the performance of various LLMs in patent claim generation. The results demonstrated that description-based claim generation outperformed previous research, which relied on abstracts \citep{jiang2024can}. Moreover, fine-tuning can enhance the claims' completeness, conceptual clarity, and logical linkage. GPT-4 achieved the best performance among all tested LLMs and uniquely grouped alternative embodiments within dependent claims logically. Despite promising capabilities, comprehensive revisions are still necessary for LLM-generated claims to pass rigorous patent scrutiny. Follow-up work further explored the ability of LLMs to revise patent claims. The findings indicate that LLMs often introduce ineffective edits that deviate from the intended revisions, whereas fine-tuning can improve performance \citep{jiang2024patent}. Conversely, \citet{wang2024patentformer} appear to be the first to investigate the generation of patent specifications from claim and drawing texts. They fine-tuned GPT-J \citep{mesh-transformer-jax} and T-5 \citep{raffel2020exploring} and demonstrated their ability to produce human-like patent specifications that adhere to a legal writing style. However, their approach still relies on certain simplifying assumptions that deviate from real-world patent drafting practices, such as the assumption that each specification paragraph corresponds to only one claim feature and a single drawing. While their work lays a solid foundation for specification generation, significant challenges remain before achieving practical applicability.

In addition, \citet{christofidellis2022pgt} proposed a prompt-based generative transformer for the patent domain, which used GPT-2 as backbone and trained with multi-task learning \citep{maurer2016benefit} on part-of-patent generation, text infilling, and patent coherence evaluation tasks. Specifically, the training processes involved two text generation tasks, generating patent titles based on abstracts and suggesting words for masked tokens in the given abstracts. The performance was better than the baseline, specifically BERT and GPT-2 on a single task. This paper indicates that multi-task learning may be a promising method for patent text generation in the future. A recent report focused on the drafting of patent sections from academic papers, such as background, summary, and description \citep{knappich2024pap2pat}. The authors highlighted the potential of LLMs in patent drafting and revealed the main challenges, including handling content repetition, retrieving contextually relevant data, and adapting the system for longer documents.

Recent studies have explored the application of LLM-based agents in various aspects of patent drafting, analysis, and management. An agent in this context refers to an autonomous system that leverages LLMs to perform specialized tasks, often in a structured, multi-agent framework where different agents collaborate to achieve complex objectives. For example, \citet{wang2024evopat} proposed EvoPat, a multi-LLM-based agent designed for patent summarization and analysis. \citet{wang2024autopatent} introduced AutoPatent, a multi-agent framework comprising an LLM-based planner agent, writer agents, and an examiner agent, which work together for patent application drafting. Additionally, \citet{chu2024paris} presented an LLM-based recommender system specifically designed to assist with patent office action responses. PatExpert \citep{srinivas2024towards} in turn suggested a meta-agent that orchestrates task-specific expert agents for various patent-related tasks, such as classification, claim generation, and summarization. It also includes a critique agent for error handling and feedback provision to enhance adaptability and reliability.

While LLMs have shown outstanding performance on generation tasks, their application in the patent domain is under-explored. Future work needs comprehensive evaluations of LLMs on patent generation tasks. The highly specialized and technical nature of patent texts may pose significant challenges for applying general LLMs to patent tasks.

\section{Future Research Directions}
\label{futurework}
\subsection{Data and Benchmarks}
Machine-learning models for the patent field, particularly LLMs, require extensive pre-structured and pre-processed high-quality data for training, testing, and knowledge extraction. Although patent databases contain vast amounts of raw data accumulated over the years, there is a lack of labeled patent datasets for different tasks. We summarize existing curated patent data collections in Section~\ref{curateddata}, showing that the number of publicly accessible datasets for some tasks is limited, such as patent novelty prediction and patent text simplification. To stimulate the development of better methods, patent offices might consider providing more of their internal pre-processed databases.

Additionally, benchmarks consist of labeled datasets and established metrics for performance evaluation, such as accuracy and precision. These benchmarks serve as a foundational framework that ensures researchers can assess the strengths and limitations of their methods against a standardized set of conditions to promote transparency and fairness in comparison. In the context of the patent domain, many tasks have relied on closed-source datasets, which limits the ability of other researchers to replicate and validate findings. Moreover, general text evaluation metrics may not be suitable for patent text assessment \citep{jiang2024can}. The absence of well-defined benchmarks for these tasks renders it challenging to assess and compare the effectiveness of various models fairly and comprehensively.

\subsection{Use of Large Language Models}

Since texts are the key ingredient to patent documents, promising methods to analyze and generate patent texts are worth investigating. Large language models (LLMs) have dramatically changed the field and achieved remarkable performance in a wide range of tasks in the general domain \citep{min2023recent}, such as information retrieval, translation, and summarization. However, few works have evaluated and investigated the usage of LLMs on patent-related tasks, leaving a large research gap. 
Patents include a distinct category of text, characterized by their formal language, dense technical content, and legal implications. The exploitation of LLMs to handle the intricacies of patent texts could yield considerable benefits in understanding and managing intellectual property.
LLMs could identify more intricate and nuanced patterns and relationships within patent texts, facilitating the effectiveness in patent analysis tasks, such as patent classification and novelty prediction. 
Furthermore, LLMs could assist in generating and refining patent texts based on a comprehensive understanding of existing patent literature. For example, LLMs could assist inventors and patent attorneys in drafting initial claims, expanding descriptions, or adapting texts to align with jurisdictional requirements.
The integration of LLMs into patent tasks cannot only enhance the efficiency and effectiveness of patent analysis but also improve the quality of the patent drafting process. 
As these models continue to evolve, LLMs appear significantly promising in the patent domain.

\subsection{Long Sequence Modeling}

Most patent analysis tasks are based on short texts, such as titles, claims, and abstracts, ignoring the patent description. One of the potential reasons is that the average length of patent descriptions far exceeds the context length of many transformer models. It is difficult to deal with such long-range dependencies and context retention. Whereas patent titles and abstracts are usually generic and vain, patent descriptions have to provide details of the invention and disclose the invention in all aspects. Hence, integrating detailed descriptions for patent analysis tasks may significantly improve performance. Therefore, a closer investigation of long-sequence modeling approaches appears to lead to a promising tool for patent descriptions. It is worth noting that the latest LLMs can support longer inputs with currently more than 100,000 tokens context length as the benchmark. As discussed in Section~\ref{s:insight}, LLMs would unlock the potential of using the patent description part.

\subsection{Patent Text Generation}

Patent text generation tasks are much less investigated, probably due to the complexity and difficulty. Automated patent-text generation poses several challenges because of its highly specialized and technical nature. Firstly, patent documents require a specific style of language and use of terminology. Secondly, the language must be formal, unambiguous, and follow legal requirements to be granted. Thirdly, the language of the description and the claims must be precise, accurate, and clear. Lack of language precision, accuracy, and clarity hampers examination, risks a small scope and may provide leverage for litigation.

Although deep-learning models, especially large language models, have demonstrated previously unexpected performance in text generation in the general domain, patent text generation methods are still scarce. The development and refinement of patent text generation models are not only helpful to the patent application process but also advantageous to research in natural language processing and related domains, such as law and technology.

Furthermore, we see an important research need for evaluation metrics for patent texts. We included common text evaluation methods in Appendix~\ref{eval_tg}. However, patent texts have specific language requirements. Hence, previous evaluation metrics may not be appropriate for patent text assessment.

\subsection{Multimodal Methods}

Apart from text, patent documents themselves and their context include other useful information, such as drawings and citations. Multimodal models can combine multiple data types to improve the performance of patent analysis \citep{huang2024large}. Research explores model architectures such as CLIP \citep{radford2021learning} and Vision Transformers \citep{dosovitskiy2020image} to bridge the gap between textual descriptions and visual data to enhance patent processing. Given the only recently increased performance gains in general multimodal methods, there are only few studies for patent applications yet, for example in patent classification \citep{jiang2022deep} and patent image retrieval \citep{pustu2021multimodal}. Ideally, researchers can adopt multimodal methods in other patent tasks as well. For example, multimodal models can assist in patent drafting from design figures, ensuring the textual descriptions and figures are aligned. Furthermore, multimodal methods can create more informative and intuitive visualizations of patent data, providing a clearer understanding of the patent landscape. A multimodal approach appears promising and deserves more attention.

\section{Conclusion}
\label{conclusion}

Patent texts are different from mundane texts in various aspects. They must be precise to ensure the patent is grantable, defensible, and enforceable. The topic, on the other hand, refers to intricate technical aspects.
The requirement for precision in combination with technical complexity has led to a highly artificial language. Reading and correct interpretation of those texts require a high level of concentration, specific patent training, and experience. Whereas patent tasks are highly manual and the knowledge contained in the patent literature is widely unused, NLP appears as an ideal solution. It can deal with complicated structures and learn definitions (terms) that deviate from every-day language or even the vocabulary of experts in the specific field.  

As NLP techniques critically rely on high-quality training data, we collected patent data sources and databases, with curated datasets specifically designed for different patent tasks. Although patent offices have provided raw patent documents for years, publicly accessible curated datasets for specific tasks are limited. Offices only offer manual reading of individual documents, instead of pre-processed data or large-scale access to well-structured legal process documents of the file (register). The field needs high-quality data for training to optimize the model performance. In addition, the proposal of novel patent tasks should suggest open-sourced datasets and benchmarks for future research.

NLP techniques play a prominent role in the automation of patent processing tasks.
Text embeddings can extract word and semantic information for patent analysis. Sentence/paragraph embeddings are preferable to word embeddings because they can capture more nuanced contextual information from entire sentences or paragraphs. From the model perspective, most traditional machine-learning models, such as feature-based neural networks, support vector machines, or random forests, have already been outdated. Deep learning models, particularly convolutional and recurrent neural networks, are still used in some circumstances, for example when the computing power is limited. However, larger transformer-based models, such as the BERT and the GPT series, have almost revolutionized the field despite their excessive need for computational resources. Furthermore, scaling the model size to LLMs has led to an explosion in performance on numerous general tasks. 

Research has prominently focused on short text parts of patents, such as titles and abstracts, e.g., for patent analysis or generation tasks. However, these texts are the least specific texts with little information about the actual invention. Since they are highly generic and take little time during drafting, the automation of these short texts is not necessarily helpful. Among the shorter parts of patents, the claims would be an exception, defining the legal scope of patent protection. The tasks of claim generation, description generation, and LLM-based patent agents are worth investigating. 

The most studied and developed topics are automated patent classification and retrieval, which are routine tasks in patent offices. Transformer-based language models have shown better effectiveness in leveraging text information compared to traditional machine learning models and deep word embeddings. A more promising direction is to hybridize methods so that they can use multiple sources from patents, including texts, images, and metadata. Information extraction serves as a primer to support various patent-related applications. State-of-the-art language models can help in precise extraction to improve the performance in further applications. Furthermore, we found that the conception of novelty used in research can substantially deviate from its strict definition in patent law. Accordingly, a share of the studies may aim at predicting the patent novelty but actually generate some form of metric for oddness, potential creativity, or perceived originality. Such cases indicate the often underestimated difficulties when working with patents. Interestingly, practically no capable foundation models can deal with the particularities of the patent domain and simplify the derivation of usable tools for specific tasks. Granting prediction, litigation prediction, and patent valuation are comparably less investigated. The key acts in patent examination are almost ignored so far, including the automated formal novelty and inventiveness detection for higher quality and objectivity, deriving of arguments, and support in formulating motions.

The exploitation of the knowledge base in the patent literature appears more developed. Technology forecasting and innovation recommendation are pattern search tasks that do not necessarily need recent language processing techniques but can already produce reasonable output with old text-parsing techniques in combination with conventional statistics and machine-learning methods. Large language models would cause an overwhelming computational burden if they had to process such large bodies of text. Thus, these fields depend more on good problem definitions and mapping to conventional pattern-identification tasks than on the latest developments of machine-learning techniques.

Latest language models play a role in generative tasks around patents and reflect the most recent trend towards generative artificial intelligence technology. 
Apart from the patent analysis tasks, the language particularities of patents also complicate generative tasks. Summarization tasks and automated translation tasks are well-established and available online. Particularly, automated translation has been studied and in production at patent offices for more than a decade. 
Less conspicuous yet socially beneficial tasks, such as patent simplification, receive scant attention in research. This oversight is primarily attributed to the absence of appropriate datasets. Although patent drafting, especially the sections on descriptions and claims, is an obvious task, the available research is limited and generally unsatisfactory.

The primary obstacles to advancing more effective techniques are twofold. First, there is a need for better access and classification of data for training and analysis. Patent laws in most countries mandate the publication of patent applications. Historically, this was considered a contract between inventors and society, offering inventors a limited monopoly in exchange for disclosing their inventions in sufficient detail to permit replication. However, the current mode of publication presents PDF documents with text extraction and related registry documents that are rudimentarily classified, often accompanied by poor-quality scans, with restrictions on automated processing. However, the necessary information would be available in these offices. Second, the field would greatly benefit from foundational models capable of handling the high degree of formality in patent language, where challenges are most prominently observed.


\bibliography{anthology,custom}

\appendix

\section{Patent Drawings and Metadata}
\label{drawings}

Apart from patent texts, researchers also use metadata and images for patent analysis. Metadata typically includes citations/references, classification codes, inventors, applicants, assignees, law firms, and examiners. Researchers have exploited the relationship information between patents through citations and classification codes based on graph neural networks (GNNs) for patent classification and patent valuation. Solely relying on classification codes can sometimes be misleading and become a problem because patents may span multiple technology areas. Furthermore, the publicly used classification is rather coarse in light of the high number of documents and notably coarser than what offices use internally for their search or assigning competent examiners.  Importantly, metadata provide high-level information but do typically not include details of the invention, which limits the possibility of capturing the invention and certainly not technical nuances. 

The images, called drawings or figures in patents, could leverage the development of image processing techniques, particularly convolutional neural networks (CNNs). However, the figures can be very generic, particularly in method inventions. The fact that patent figures are not supposed to contain (significant) text beyond reference numbers, can have technological advantages. Patent figures from computer aided design (CAD) models may be harder to match with corresponding ones from other documents due to the high number of lines representing details, which tend to overwhelm convolutional neural networks. However, many patents and patent applications contain simpler sketches and schematic figures with the intention to not disclosing more than necessary. Such simplifications can be beneficial for processing with artificial intelligence. By now, researchers have applied convolutional neural networks in patent subject classification and patent retrieval. However, considering the above, the performance of patent classification based on images only appears to be much worse than using texts. Therefore, hybrid methods that leverage multiple sources for patent analysis, such as texts, images, and classification codes, appear promising.

The drawings are not helpful for automated processing in many cases. Methods are commonly illustrated as block diagrams, but these diagrams, without supplementary text beyond reference numbers—which many patent offices do not accept for formal reasons—become ambiguous and could be interpreted as representing completely different and undeniably novel inventions. Furthermore, the size and rapid growth of patent descriptions (including applications that are not granted) and the scientific-technical literature worldwide pose significant challenges to the prior-art search.

\subsection{Patent Classification}
\label{a:classification}
Images can serve for automated patent classification \citep{jiang2021deriving}. There is usually no specific feature extraction process. Deep-learning networks, typically convolutional neural networks, can automatically extract image features from the pixel values. Jiang et al.\ designed a method based on convolutional neural networks for patent classification using only images and achieved 54.32\% accuracy on eight-class prediction \citep{jiang2021deriving}. The results are poorer than text-based methods for two main reasons. Firstly, predicting the subject class based solely on images is inherently challenging, even for human experts\footnote{Important background information: in most patent offices, the drawings are not supposed to contain text beyond reference numbers.}. Secondly, some patent images suffer from low resolution. However, using images as additional information for patent classification on top of the text may be a promising research direction.

\subsection{Patent Retrieval}
\label{a:retrieval}
\textbf{Metadata-Based Methods. } Commonly used metadata includes citations\,{}/\,{}references, classification codes, inventors, applicants, assignees, and examiners.  Involving metadata for patent retrieval may improve the retrieval performance because metadata can imply relevance among documents \citep{shalaby2019patent}. For example, citations naturally indicate the relationship between patent documents, documents that belong to the same category may have connections, and an inventor may tend to investigate similar types of inventions. A representative study would be the embedding method that used both texts and metadata for patent representation \citep{siddharth2022enhancing}. The authors aggregated text embeddings obtained from Sentence-BERT and citation embedding obtained from knowledge graphs to formulate the patent representation. This method improved the accuracy by approximately 6\% compared to using text only.

\noindent\textbf{Image-Based Methods. } Recently, the rapid development of image processing techniques has directed the attention of retrieval tasks to patent drawings. \cite{kucer2022deeppatent} introduced a large-scale dataset, DeepPatent, for an image-based patent retrieval benchmark. The dataset includes 45,000 unique design patents and 350,000 drawings from 2018 and the first half of 2018 from the United States Patent and Trademark Office. This paper also implemented the PatentNet model as a baseline, a deep learning approach based on the ResNet architecture \citep{he2016deep}. It achieved a 0.376 mean average precision (MAP), which was better than traditional computer vision approaches and other deep representations tested in their experiments. After that, \cite{wang2023learning} improved the performance to 0.712 by leveraging the unique characteristics of patent drawings. The authors used EfficientNet-B0 \citep{tan2019efficientnet} as the backbone, which is an effective and efficient convolutional network. They also incorporated a neck structure that comprises a fully connected layer and a consecutive batch normalization layer to improve the intra-class compactness and inter-class discrimination. In addition, \cite{higuchi2023patent} proposed a transformer-based deep metric learning architecture, which reached a new state-of-the-art mean average precision score of 0.856 on the DeepPatent dataset. 

\subsection{Figure Caption/Description Generation}

Researchers studied generating brief descriptions of the patent figures. \citet{aubakirova2023patfig} introduced the first large-scale patent figure dataset, comprising more than 30,000 figure--caption pairs. The authors fine-tuned large vision language models to generate captions as baselines on this novel dataset and call for future research on the improvement of existing caption generation models. As contextual information, the quality of such brief figure descriptions can vary significantly. In quite many cases, the figure captions read \textit{Figure n depicts another embodiment of the invention} or \textit{Figure m depicts perspective xy of an embodiment of the invention} to fulfill the minimum formalities (e.g., per European Patent Regulations Rule 42(d), PCT Regulations Rule 5.1(a)(iv), US 37 CFR 1.74) and let the subsequent detailed description of the figures and embodiments provide more content. Similarly, \citet{shukla2025patentlmm} presented PatentDesc, a novel large-scale dataset that comprises approximately 355K patent figures paired with both brief and detailed textual descriptions. They trained a vision encoder specifically designed for patent figures and demonstrated that it can significantly enhance the generation performance.
Furthermore, \citet{shomeeimpact} introduced IMPACT, a multimodal patent dataset that contains 3.61 million design patent figures paired with detailed captions and rich metadata. They used the dataset to build benchmarks for two novel computer vision tasks---3D image construction and visual question answering (VQA).

\section{Evaluation Metrics}
\label{evaluationmetric}
Evaluation metrics quantify a model's performance, offering a clear and objective measure of how well the model can achieve the intended task. This is essential for understanding the model's capabilities and limitations. In addition, metrics allow researchers and developers to compare the effectiveness of different models or algorithms on the same task. By understanding the most relevant metrics to a specific task, developers can select or design models that are optimized for those metrics. Different metrics are used based on task types. We introduce some common evaluation metrics for different tasks, including classification (Section~\ref{eval_classification}), information retrieval (Section~\ref{eval_ir}), and text generation (Section~\ref{eval_tg}). 

\subsection{Classification}
\label{eval_classification}
Widely used metrics for classification tasks are \textbf{accuracy (A)}
\begin{equation}
    A=\frac{TP+TN}{TP+TN+FP+FN},
\end{equation}
\textbf{precision (P)}
\begin{equation}
    P=\frac{TP}{TP+FP}, 
\end{equation}
\textbf{recall (R)}
\begin{equation}
    R=\frac{TP}{TP+FN},
\end{equation}
and \textbf{F1 scores (F1)}
\begin{equation}
    F1=\frac{2PR}{P+R},
\end{equation}
where true positives (TP) refer to the number of samples that are correctly predicted as positive, true negatives (TN) the number of samples that are correctly predicted as negative, false positives (FP) the number of samples that are wrongly predicted as positive, and false negatives (FN) the number of samples that are wrongly predicted as negative.

There are two methods of aggregating performance metrics across classes, namely macro-averaging and micro-averaging. \textbf{Macro-averaging} calculates the metric independently for each class and then takes the numerical average. This means the weights of all classes are equal, regardless of size or frequency. For example, macro-averaged precision can be calculated per
\begin{equation}
Macro Precision = \frac{1}{N} \sum_{i=1}^{N} P_i,
\end{equation}
where $N$ is the number of classes. 

In \textbf{micro-averaging}, the process involves summing up all TP, FP, and FN respectively across classes, followed by the calculation of the metric. By default, micro-averaging often serves as the preferred method unless specified otherwise, due to its ability to more appropriately handle class imbalances by assigning more weight to the majority class. Micro-averaged precision can for instance follow
\begin{equation}
Micro Precision = \frac{\sum_{i=1}^{N} (TP)_i}{\sum_{i=1}^{N} ((TP)_i + (FP)_i)}.
\end{equation}

\subsection{Information Retrieval}
\label{eval_ir}
In information retrieval tasks, \textbf{precision (P)}
\begin{equation}
    P=\frac{N_R}{N_T}, 
\end{equation}
\textbf{recall (R)}
\begin{equation}
    R=\frac{N_R}{N_K},
\end{equation}
and \textbf{mean average precision (MAP)}
\begin{equation}
MAP = \frac{1}{Q} \sum_{q=1}^{Q} \left( \frac{1}{m_q} \sum_{k=1}^{n_q} P(k) \cdot \sigma(k) \right)
\end{equation}
usually serve for evaluation \citep{baeza1999modern}. 

$N_R$ is the number of relevant documents retrieved, $N_T$ the total number of documents retrieved, and $N_K$ the total number of really relevant documents.  $Q$ denotes the total number of required queries, $m_q$ the number of relevant documents for the \textit{q-th} query, $n_q$ the number of documents retrieved for the \textit{q-th} query, $P(k)$ the precision at the \textit{k-th} position, and $\sigma(k)$ an indicator function that equals 1 if the \textit{k-th} document is relevant, and 0 otherwise.

In addition, the \textbf{patent retrieval evaluation score (PRES)}
\begin{equation}
PRES = 1 - \frac{\frac{\sum_{i=1}^{N} r_i}{N} - \frac{N+1}{2}}{N_{max} }
\end{equation}
 is a metric specifically designed to evaluate the performance of patent retrieval systems \citep{magdy2010pres}. 
$N$ represents the number of relevant documents, $N_{max}$ the maximum number of documents returned by the retrieval system, and $r_i$ the rank of the \textit{i-th} relevant document in the retrieval results. It calculates the average deviation of the actual ranking of relevant documents from their ideal ranking with normalization.

\subsection{Text Generation}
\label{eval_tg}
Evaluation metrics provide feedback that is used to adjust the model's parameters and improve its performance. This optimization is critical to developing highly accurate and efficient models. The most reliable but expensive and slow evaluation methods for text generation are human evaluations. \textbf{Human evaluation} involves assessing the quality of the generated text by human experts. Multiple aspects are evaluated simultaneously, such as accuracy, fluency, coherence, and relevance. Human evaluation is often considered the gold standard, but it can be time-consuming and expensive.

Researchers have developed automated evaluation methods for text generation to improve efficiency and reduce costs. These approaches compare generated texts with referenced texts (gold standard) to obtain the performance score. We briefly introduce four of the most commonly used methods, including BLEU \citep{papineni2002bleu}, ROUGE \citep{lin2004rouge}, METEOR \citep{banerjee2005meteor}, and BERTScore \citep{zhang2019bertscore}. 

\textbf{Bilingual evaluation understudy (BLEU)} score \citep{papineni2002bleu} quantifies how much the generated texts match the high-quality reference texts, by comparing the n-grams to count the number of exact matches. BLEU is simple and easy to use, but it only focuses on literal matches, overlooking overall sentence semantics.  BLEU was originally designed for machine translation but could also be used for other generation tasks. 

\textbf{Recall-oriented understudy for gisting evaluation (ROUGE)}  counts the number of overlapping units, such as n-gram and word sequences, to obtain the performance score \citep{lin2004rouge}. While BLEU focuses more on precision, i.e., how many generated texts match reference texts, ROUGE prioritizes more recall performance, i.e., how many reference texts are covered by generated texts. Hence, this method is more suitable for text summarization tasks. Similar to BLEU, it evaluates text content, without considering semantics or synonyms.

\textbf{Metric for evaluation of translation with explicit ordering (METEOR)} improves on BLEU by accounting for both exact word matches and similar words based on stemming and synonyms \citep{banerjee2005meteor}. Thus, METEOR can better account for semantic variations compared to BLEU. In addition, METEOR provides a more balanced assessment by considering both precision and recall. However, METEOR is more complex and needs adjustments for different tasks. 

\textbf{BERTScore} \citep{zhang2019bertscore} leverages the contextual embedding from BERT \citep{devlin2018bert} and calculates cosine similarity between output text embedding and referenced embedding. It can effectively capture complex and subtle semantic information but requires significant computational resources for evaluation.

Recently, \citet{lee2023evaluating} proposed a novel \textbf{keystroke-based evaluation} method for generative patent models. This method measured the number of keystrokes that the model could save by providing model predictions in an auto-complete function. The experimental results demonstrated that the larger models could not improve this specific evaluation metric. However, the experiments only focused on patent claims generation and the effectiveness of this metric on other generative tasks was unknown. 

A recent work proposed \textbf{PatentEval} as a human evaluation framework to evaluate machine-generated patent texts \citep{zuo2024patenteval}. The framework categorizes errors into different types, such as grammatical errors, irrelevant content, incomplete coverage, and clarity issues. The authors manually assessed 400 granted patents and indicated that while LLMs demonstrate potential for patent drafting, they still exhibit limitations.

\end{document}